%% file: example_paper.tex
\documentclass{article}

\usepackage{microtype}
\usepackage{graphicx}
\usepackage{subcaption}
\usepackage{booktabs} % for professional tables

\usepackage{hyperref}

\usepackage[preprint]{icml2026}

\usepackage{amsmath}
\usepackage{amssymb}
\usepackage{mathtools}
\usepackage{amsthm}
\usepackage{graphicx}
\usepackage{algorithm}
\usepackage{algpseudocode}
\usepackage{listings}         
\usepackage{xcolor}
\usepackage{booktabs}
\usepackage{enumitem}

\usepackage{multirow}
\usepackage[table,xcdraw]{xcolor}
\usepackage{hhline}     
\usepackage{makecell}   
\definecolor{tableheadcolor}{gray}{0.9} 
\usepackage{textcomp}
\usepackage{colortbl}

\usepackage[capitalize,noabbrev]{cleveref}

\theoremstyle{plain}

\theoremstyle{definition}

\theoremstyle{remark}

\usepackage[textsize=tiny]{todonotes}

\icmltitlerunning{Towards Efficient and Scalable Neural Graph Databases Training}

\begin{document}

\twocolumn[
  \icmltitle{NGDB-Zoo: Towards Efficient and Scalable Neural Graph Databases Training}
  % \icmltitle{NGDB-Zoo: Scaling Neural Graph Databases with Decoupled Semantic Priors}



  \icmlsetsymbol{equal}{*}

  \begin{icmlauthorlist}
    \icmlauthor{Zhongwei XIE}{yyy}
    \icmlauthor{Jiaxin BAI}{yyy}
    \icmlauthor{Shujie LIU}{comp}
    \icmlauthor{Haoyu HUANG}{yyy}
    \icmlauthor{Yufei LI}{yyy}
    \icmlauthor{Yisen GAO}{yyy}
    \icmlauthor{Hong Ting TSANG}{yyy}
    \icmlauthor{Yangqiu SONG}{yyy}
  \end{icmlauthorlist}

  \icmlaffiliation{yyy}{The Hong Kong University of Science and Technology}
  \icmlaffiliation{comp}{Microsoft Research Asia}

  \icmlcorrespondingauthor{Yangqiu SONG}{yqsong@cse.ust.hk}


  \vskip 0.3in
]

\newcommand{\up}[1]{\textsuperscript{\textcolor{gain}{$\uparrow$#1}}}
\newcommand{\down}[1]{\textsuperscript{\textcolor{loss}{$\downarrow$#1}}}



\printAffiliationsAndNotice{}  % no special notice (required even if empty)

\newcommand{\method}{{NGDB-Zoo}}

\newcommand{\shortmethod}{{NGDB-Zoo}}

\begin{abstract}
Neural Graph Databases (NGDBs) facilitate complex logical reasoning over incomplete knowledge structures, yet their training efficiency and expressivity are constrained by rigid query-level batching and structure-exclusive embeddings. We present \textbf{\method{}}, a unified framework that resolves these bottlenecks by synergizing \textbf{operator-level training} with \textbf{semantic augmentation}. By decoupling logical operators from query topologies, \method{} transforms the training loop into a dynamically scheduled data-flow execution, enabling multi-stream parallelism and achieving a $1.8\times$ - $6.8\times$ throughput compared to baselines. Furthermore, we formalize a decoupled architecture to integrate high-dimensional semantic priors from Pre-trained Text Encoders (PTEs) without triggering I/O stalls or memory overflows. Extensive evaluations on six benchmarks, including massive graphs like \textit{ogbl-wikikg2} and \textit{ATLAS-Wiki}, demonstrate that \method{} maintains high GPU utilization across diverse logical patterns and significantly mitigates representation friction in hybrid neuro-symbolic reasoning. 
\end{abstract}

\section{Introduction}
% NGDB is a “neuralized” storage engine that, by integrating traditional symbolic representations (nodes and edges) with deep learning methods like vector embeddings, has become an important advancement in the field of Neural-Symbolic AI.

Neural Graph Databases (NGDBs) have emerged as a powerful paradigm for complex logical reasoning over large-scale, incomplete knowledge structures \cite{ren2023neural, besta2022neural,shi2025accurate,pengmei2024pushing}. By embedding existential first-order logic queries into continuous geometric spaces, these systems enable approximate retrieval and inductive inference beyond symbolic traversal \cite{ren2020query2box, bai2023sequential}. Despite their promise, the transition from theoretical models to production-ready query engines is hindered by a fundamental tension between \textbf{hardware efficiency} and \textbf{representational expressivity}.

We identify two prohibitive challenges that prevent the realization of scalable NGDBs:
\textbf{First, Computational Efficiency and Topological Rigidity.}
Existing dense-retrieval-based training frameworks rely on batching at the query level, which constrains a single mini-batch to queries sharing isomorphic logical structures (e.g., 2i) \cite{ren2020beta,ramezanpour2025few,zeng2024phympgn}. 
% Existing training frameworks rely on query-level batching, which restricts a single batch to queries with identical logical structures (e.g., 2-hop paths) \cite{ren2020beta}. 
This constraint causes severe hardware under-utilization when processing diverse query mixtures, as kernels are fragmented by the high entropy of real-world query workloads. As link density drops in large-scale graphs, this efficiency bottleneck becomes a barrier to convergence.
\textbf{Second, Representation Friction in Hybrid Integration.} While treating entities as atomic tokens is sufficient for dense Knowledge Graphs, NGDBs require richer semantic units to perform robust reasoning in sparse or inductive settings. Augmenting structural embeddings with high-dimensional semantic priors (e.g., from Pretrained Language Models) can compensate for graph sparsity~\cite{wei2025semantic,wang2021kepler,kim2022pure,chen2025autogfm,feng2025graph,tang2024training}, however it introduces significant system-level friction. The integration of large semantic vectors often triggers I/O stalls and memory overflows during multi-processing, creating a trade-off where representational gains are offset by catastrophic training latency.

\begin{figure}[t]
\centering
\includegraphics[width=0.48\textwidth]{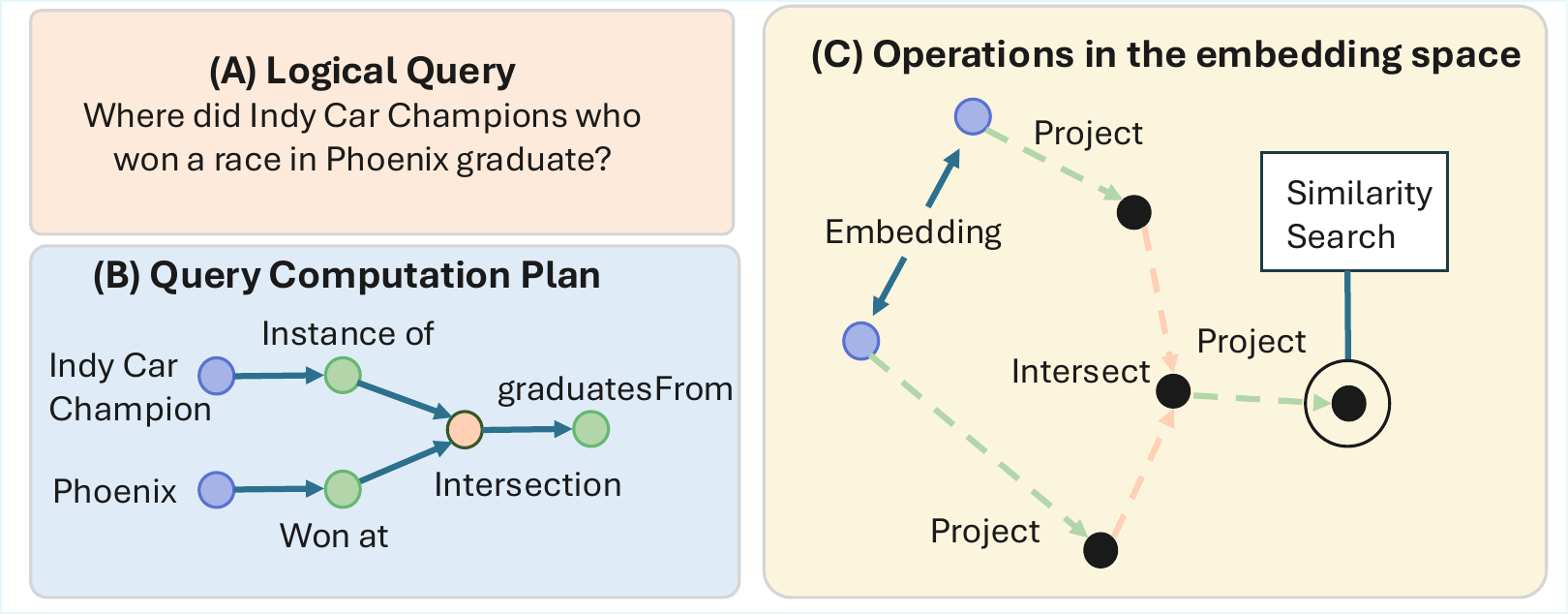}
\caption{\textbf{Task Illustration.} Query embedding methods aim to answer multihop logical queries (A) by avoiding explicit knowledge
graph traversal and executing the query directly in the embedding space by following the query computation plan (B).
Operators are executed in the embedding space (C).\label{fig:logical_query} }
    \vspace{-5mm}
\end{figure}

% \begin{figure}[t]
% \centering
% \includegraphics[width=0.48\textwidth]{fig/image_cropped.pdf}
% \caption{Conceptual comparison illustrating how \method{} overcomes the fragmented bottlenecks of training NGDB through a unified, dynamic data-flow architecture that achieves high through.\label{fig:gpu_util} }
%     \vspace{-5mm}
% \end{figure}

\noindent To resolve these challenges, we present \textbf{\method{}}, a unified framework designed for high-throughput and semantically-aware neural graph reasoning. Our approach decouples logical operators from fixed query topologies, transforming the training loop into a dynamically scheduled data-flow execution. Our contributions are as follows:\\
\noindent \textbf{Contribution 1: Operator-Level Training Paradigm.} 
We propose a novel training paradigm that shifts the batching granularity from query-level isomorphism to operator-level batching. By decomposing complex queries into a Directed Acyclic Graph (DAG) of operators, this enables dynamic scheduling and cross-query operator fusion, achieving a SOTA training throughput over standard baselines.\\
\noindent \textbf{Contribution 2: Decoupled Semantic Integration Architecture.} We develop a modal-agnostic framework to fuse structural embeddings with high-dimensional external signals from Pretrained Text Encoders (PTEs). To mitigate system bottlenecks, we implement decoupled semantic encoding and high-throughput GPU-Resident integration, enabling the efficient handling of semantic parameters without I/O blocking. \\
\noindent \textbf{Contribution 3: Scalable Neuro-Symbolic Benchmarking.} We provide a rigorous evaluation across six benchmarks, including massive graphs like \textit{ogbl-wikikg2} and \textit{ATLAS-Wiki}. Experimental results demonstrate that \method{} achieves a 1.8$\times$--6.8$\times$ increase in training throughput compared to state-of-the-art baselines. Our analysis characterizes the efficiency-expressivity trade-off in hybrid models and offers a technical roadmap for handling non-stationary query distributions through adaptive online sampling.

\begin{figure*}[t]
    \centering
    % \vspace{-3mm}
    \includegraphics[width=1.0\linewidth]{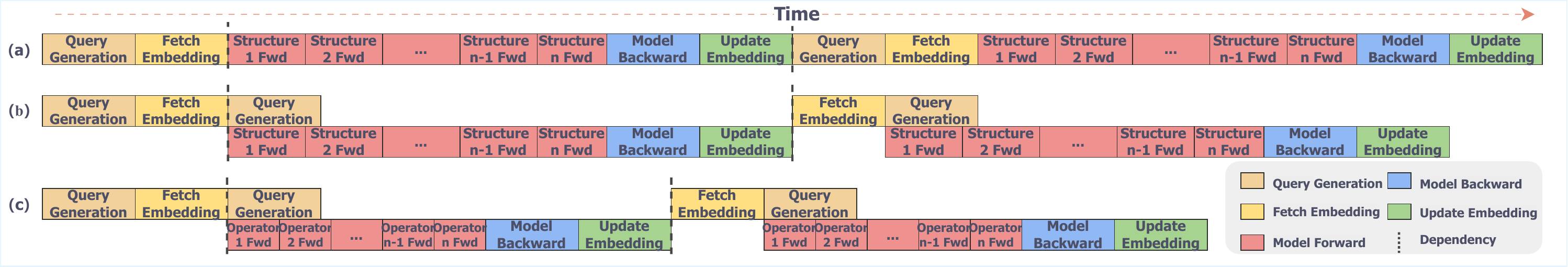}
\caption{
    \textbf{Evolution toward High-Throughput Asynchronous Pipelining.} Training methods. (a) represents the naive training scheme; (b) shows the pre-sampling and pre-fetching
optimization; (c) \method{}'s operator-level approach breaks these constraints to enable fully overlapping stages, creating a dense execution stream that maximizes hardware saturation. 
}
    \vspace{-3mm}
    \label{fig:operator_level_batching}
\end{figure*}

% Branch Contribution 3
% \noindent \textbf{Contribution 3: Difficulty-Aware Adaptive Online Sampling.} 
% Addressing the infinite query space in large-scale KGs (e.g., \textit{ATLAS-Wiki}), we introduce an online stochastic sampling system. We further propose an \textit{adaptive sampling} mechanism that dynamically reweights training workloads based on real-time difficulty scores. Experimental results demonstrate that this approach improves MRR by \textbf{16.8\%} under non-stationary query distributions and maintains consistently high GPU utilization across 14 distinct EFO logical patterns.

% We propose a novel training paradigm that decomposes complex queries into a Directed Acyclic Graph (DAG) of atomic operators and shifts the batching granularity. By shifting the batching granularity, our training paradigm transcends the conventional requirement of query-level isomorphism—which traditionally constrains a single mini-batch to queries of identical logical structures (e.g., $2i$). This framework achieves a \textbf{1.78$\times$ cumulative speedup} over standard baselines.

\section{Related Work}

\textbf{Neural Graph Databases Reasoning.} 
Complex Query Answering (CQA) extends relational link prediction to multi-hop logical operations. Early methods~\citep{yang2014embedding, trouillon2016complex} focused on simple relational patterns. More recent geometric embedding models~\citep{xiong2023geometric} represent queries as points~\citep{, sun2019rotate}, bounded regions~\citep{ren2020query2box, zhang2021cone,wu2024query2gmm}, or probabilistic distributions~\citep{ren2020beta,yang2022gammae} to capture topological constraints and uncertainty. Complementary approaches utilize hyperbolic projections~\citep{choudhary2021self} or sequential encoding~\citep{bai2023sequential}. Recently, standalone models have evolved into integrated Neural Graph Databases (NGDBs) ~\citep{besta2022neural, ren2023neural, hu2024privacy, hu2024federated,zhengtest,wang2025griffin,choi2025rdb2g}, supporting neuro-symbolic retrieval at scale~ and providing the state-space foundation for autonomous planning in Agentic NGDBs~\citep{bai2025top,lefebvre2025scalable}. Despite their promise, the infinite, structurally diverse query space in NGDBs renders traditional offline training prohibitive, necessitating the high-throughput online sampling and operator-level optimization proposed in this work.

\noindent\textbf{Training Optimization for Graph Embedding Models.}
% Standard training on graphs regimes rely on isomorphic batching~\citep{}.
Graph Neural Networks~\citep{wu2020comprehensive,galkin2020message,hegoing,finder2025improving,liyou,bao2025beyond} use neighbor sampling or subgraph batching. Sequence Models~\citep{ott2019fairseq, yan2021fastseq} batch sequences with similar lengths, padding when necessary. Knowledge Graph Embeddings~\citep{ren2022smore,ge2024knowledge,cuiimproving} batch simple link prediction triples.
Prior work has explored operator-level parallelism in different contexts. Compiler Optimization~\citep{chen2023dycl} batches operators in computational graphs for inference. Data Flow Architectures~\citep{veen1986dataflow} schedule operations based on data availability. Atom~\citep{zhou2024atom} introduced operator-level batching to optimize throughput in KG serving and inference systems.
To the best of our knowledge, \method{} is the first to apply operator-level batching to the training of knowledge graph embedding models.

% \noindent \textbf{Operator-Level Batching.} Prior work has explored operator-level parallelism in different contexts, such as DNN-based video analysis~\citep{shen2019nexus}, Transformer-based generative models~\citep{yu2022orca}, and online prediction serving~\citep{crankshaw2017clipper}. \textbf{Compiler Optimization}~\citep{chen2023dycl} batches operators in computational graphs for inference. Data Flow Architectures~\citep{veen1986dataflow} schedule operations based on data availability. Atom~\citep{zhou2024atom} introduced operator-level batching to optimize throughput in KG serving and inference systems.

\textbf{Pretrained Embeddings for Structured Knowledge.}
Besides dense-retrieval-based method, recent work has explored combining pretrained language models with knowledge graphs~\citep{ketata2025joint,yao2019kg}: KEPLER~\citep{wang2021kepler} leverages KG embeddings as supervision signals to train language models. SimKGC~\citep{wang2022simkgc} uses contrastive learning to align entity descriptions with KG structure. Some works~\citep{galkin2023towards, galkin2024foundation, yun2019graph,min2022transformer, hu2020heterogeneous, rong2020self,chen2022structure,chen2025flatten,kim2024discrete} build pre-trained GNN foundation model for reasoning tasks on KGs. 
Recent state-of-the-art encoders~\citep{wang2021tsdae} like Qwen3-Embedding~\citep{qwen3embedding} and BGE-Embedding~\citep{bge_embedding} provide rich semantic similarity signals. 
However, integrating high-dimensional PLM outputs into complex multi-hop reasoning often triggers severe I/O stalls and memory bottlenecks. 
Fusing these semantic representations with multi-hop reasoning models efficiently remains an open challenge.

\section{Problem Formulation}
\label{sec:problem_formulation}

\subsection{Complex Logical Queries}
Let $\mathcal{G} = (\mathcal{E}, \mathcal{R}, \mathcal{T})$ be a knowledge graph, where $\mathcal{E}$ is the set of entities, $\mathcal{R}$ is the set of relations, and $\mathcal{T} \subseteq \mathcal{E} \times \mathcal{R} \times \mathcal{E}$ represents the observed fact triples. We consider the task of answering complex logical queries $q$ that can be expressed in Existential First-Order (EFO) logic, following the form:
\begin{equation}
    q[V_?] = \exists V_1, V_2, \dots, V_k : \phi(V_?, V_1, \dots, V_k)
\end{equation}
where $V_?$ is the target variable (the answer), $\{V_1, \dots, V_k\}$ are existentially quantified bound variables, and $\phi$ is a conjunction or disjunction of atomic formulas (literals) such as $(V_a, r, V_b)$. 

Neural Graph Databases (NGDBs) execute complex logical queries over these structures. We evaluate 14 distinct query patterns involving projection ($p$), intersection ($i$), union ($u$), and negation ($n$). These patterns ($1p$, $2p$, $3p$, $2i$, $3i$, $pi$, $ip$, $2u$, $up$, $2in$, $3in$, $pin$, $pni$, $inp$) represent different topological constraints in the reasoning subgraph, challenging the model to perform multi-hop compositional reasoning.

\subsection{The Task: Predictive Query Answering}
\label{sec:task_qa}

Unlike standard database queries that rely on a complete $\mathcal{G}$, the \textbf{Predictive Query Answering} task requires identifying answers from an incomplete graph where many true triples are missing during the training and inference time.

\paragraph{Answer Set.} 
For a given query $q$, we define its \textit{denotation set} $\mathcal{A}_q \subseteq \mathcal{E}$ as the complete set of entities satisfying the logical constraints of $q$ under the ground-truth graph $\mathcal{G}_{full}$. While our framework optimizes a unified objective over $\mathcal{A}_q$, this set is implicitly comprised of two conceptually distinct categories: \textit{Direct Answers} ($\mathcal{A}_q^{obs}$), which are reachable via symbolic execution on the observed training graph $\mathcal{G}_{train}$, and \textit{Predictive Answers} ($\mathcal{A}_q^{miss} = \mathcal{A}_q \setminus \mathcal{A}_q^{obs}$), which are valid under $\mathcal{G}_{full}$ but remain latent in $\mathcal{G}_{train}$ due to missing triples. This implicit distinction allows us to evaluate the framework's capacity to generalize from observed relational patterns to inductive reasoning over missing knowledge.

\paragraph{Objective Function.}
The objective is to learn a scoring function $f: \mathcal{Q} \times \mathcal{E} \to \mathbb{R}$, where $\mathcal{Q}$ is the query space. For each $e \in \mathcal{E}$, $f(q, e)$ measures the likelihood that $e$ is a valid answer to $q$. The task is formulated as a ranking problem:
\begin{equation}
    \forall e \in \mathcal{A}_q, e' \notin \mathcal{A}_q \implies f(q, e) > f(q, e')
\end{equation}
The model must generalize beyond memorized paths by embedding the query into a latent representation $\mathbf{q} \in \mathcal{H}_{joint}$ (as defined in Section~\ref{sec:joint_formulation}) and computing its proximity to entity embeddings $\mathbf{x}_i$ in the joint semantic-structural space.

\paragraph{Joint Structural-Semantic Representation.}
We define a joint representation setting where the entity embedding space is a composite manifold $\mathcal{H}_{joint} = \mathcal{H}_{str} \times \mathcal{H}_{sem}$. The structural subspace $\mathcal{H}_{str} \subseteq \mathbb{R}^{d_s}$ captures endogenous graph topology through trainable embeddings $\mathbf{h}_i$, while the semantic subspace $\mathcal{H}_{sem} \subseteq \mathbb{R}^{d_m}$ encapsulates exogenous linguistic priors via a frozen Pre-trained Text Encoder (PTE), $\Phi$. For each entity $e_i \in \mathcal{E}$, its joint representation is $\mathbf{x}_i = [\mathbf{h}_i \oplus \Phi(d_i)] \in \mathcal{H}_{joint}$.

To integrate these features into probabilistic reasoning frameworks (e.g., BetaE), we interpret the joint signals as the sufficient statistics that parameterize entity distributions. We employ a trainable mapping network $\Psi_{\theta}: \mathcal{H}_{joint} \to \mathcal{D}$ to project the joint embeddings into the manifold $\mathcal{D}$:
\begin{equation}
    [\boldsymbol{\alpha}_i^0, \boldsymbol{\beta}_i^0] = \Psi_{\theta}(\mathbf{x}_i)
\end{equation}
where $\boldsymbol{\alpha}_i^0, \boldsymbol{\beta}_i^0$ denote the initial Beta distribution parameters. These parameters then serve as the input for the subsequent sequence of logical operators (e.g., projection, intersection), ensuring that the entire reasoning chain is grounded in both learned topology and static semantic knowledge.

% To integrate these features into diverse reasoning frameworks (e.g., GQE, Q2B, BetaE), we interpret the joint signals as the initialization parameters defining the entity's representation in the target logical manifold $\mathcal{D}$. We employ a trainable mapping network $\Psi_{\theta}: \mathcal{H}_{joint} \to \mathcal{D}$ to project the joint embeddings into the manifold $\mathcal{D}$:
% \begin{equation}
%     \mathbf{z}_i = \Psi_{\theta}(\mathbf{x}_i)
% \end{equation}
% where $\mathbf{z}_i$ denotes the model-specific initial entity representation (e.g., coordinates for vector models, center and offset for box embeddings, or sufficient statistics for probabilistic distributions). These representations then serve as the input for the subsequent sequence of logical operators, ensuring that the entire reasoning chain is grounded in both learned topology and static semantic knowledge.

\section{Method}

To address the hardware efficiency bottlenecks and representational sparsity in Neural Graph Databases (NGDBs), we propose \method. This framework decouples the logical query topology from hardware execution by decomposing complex queries into a unified pool of neural operators.

% \begin{figure*}[t]
%     \centering
%     \vspace{-3mm}
%     \includegraphics[width=1.0\linewidth]{fig/main.pdf}
% \caption{
%     \textbf{An overview of our training framework, \method{}.} 
% }
%     \vspace{-4mm}
%     \label{fig:framework_overview}
% \end{figure*}

\subsection{Operator-Level Batching}
\label{sec:operator_level_batching}

\begin{figure}[t]
    \centering
    % \vspace{-3mm}
    \includegraphics[width=1.0\linewidth]{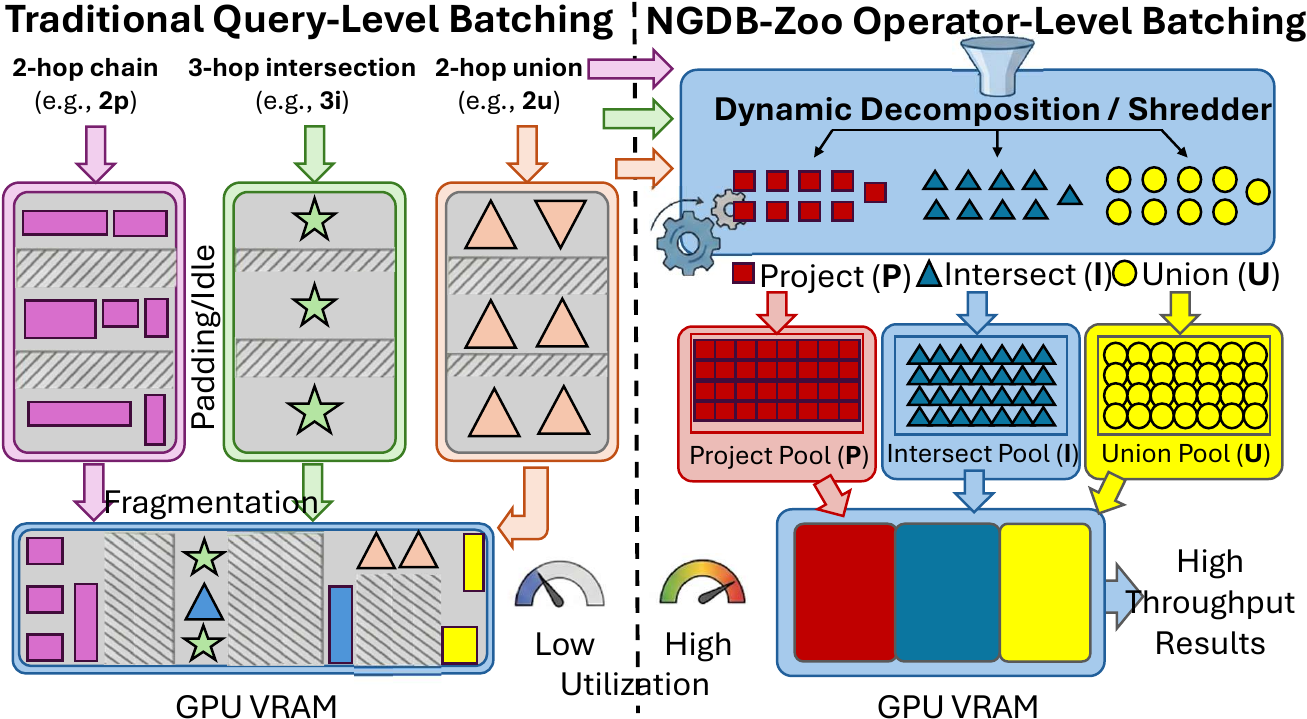}
% \caption{
%     \textbf{Paradigm shift to Operator-Level Efficiency.} 
%     \textbf{(Left):} Traditional query-level batching suffers from severe \textbf{fragmentation and padding} due to isolated processing of disparate structures. 
%     \textbf{(Right):} \method{} aggregates atomic operators into homogeneous \textbf{Operator Pools}, enabling unified kernel execution that eliminates topological constraints and maximizes GPU occupancy.
% }
\caption{
    \textbf{Overcoming Topological Rigidity.} 
    \textbf{(Left):} Query-level batching incurs high fragmentation overheads by segregating distinct query structures. 
    \textbf{(Right):} \method{} aggregates atomic operators into homogeneous Operator Pools for unified execution, eliminating structural constraints and saturating GPU cores.
}
    \vspace{-5mm}
    \label{fig:operator_level_batching}
\end{figure}

Standard query-level batching organizes computation by grouping queries with identical topological structures (e.g., grouping all 2-hop chains). However, this approach causes severe fragmentation when training on diverse query mixtures typical of Neural Agentic Databases and Neural Graph Databases. If the training distribution $\mathcal{D}$ contains $|\mathcal{T}|$ distinct query structures, the effective batch size for any specific kernel is reduced by a factor of approximately $|\mathcal{T}|$. To resolve this bottleneck, we draw inspiration from Atom~\citep{zhou2024atom}, which introduced operator-level batching to optimize throughput in KG serving and inference systems. In Figure \ref{fig:operator_level_batching}, we adapt this paradigm to the training phase of logical query embeddings. By shifting the abstraction level from the monolithic query to the atomic operator, we enable high-throughput training even the query workload is highly entropic and structurally diverse.

% Standard query-level batching organizes computation by grouping queries with identical topological structures (e.g., grouping all 2-hop chains). However, this approach causes severe fragmentation when training on diverse query mixtures; if the training distribution $\mathcal{D}$ contains $|\mathcal{T}|$ distinct query structures, the effective batch size for any specific kernel is reduced by a factor of approximately $|\mathcal{T}|^{-1}$. To resolve this bottleneck, we shift the abstraction level from the query to the atomic operator.

\paragraph{Graph Decomposition and Representation.}
We define a logical query $q$ not as a monolithic sequence, but as a directed acyclic graph (DAG) $G_q = (V, E)$. Here, $V$ represents the set of atomic operators (e.g., $\text{Project}$, $\text{Intersect}$, $\text{Union}$), and $E$ represents the data dependencies between them. Formally, a complex query such as a 2-way intersection $q = \phi_{\cap}(\phi_{p}(e_1, r_1), \phi_{p}(e_2, r_2))$ is decomposed into independent projection nodes whose outputs flow into a central intersection node.

This decomposition allows us to aggregate operators $v$ across disparate query structures into global execution pools $\mathcal{P}_\tau$, where $\tau$ denotes the operator type (e.g., matrix-vector multiplication or set intersection).

\paragraph{Dynamic Scheduling.}
\label{sec:dynamic_scheduling}
The core of our training engine is a dynamic scheduler that optimizes GPU occupancy. At any simulation step $t$, let $\mathcal{R}_t$ be the set of all ready operators (those whose dependencies are satisfied). The scheduler groups $\mathcal{R}_t$ by operator type $\tau$ and selects a target type $\tau^*$ to execute. We employ a \textit{Max-Fillness} policy to maximize hardware utilization. We define the fillness ratio $\rho(\tau)$ as:
\begin{equation}
    \rho(\tau) = \frac{|\{o \in \mathcal{R}_t : \text{type}(o) = \tau\}|}{B_{\max}}
\end{equation}
where $B_{\max}$ is the maximum efficient batch size for the hardware. The scheduler selects $\tau^* = \arg\max_\tau \rho(\tau)$, thereby prioritizing the operator type that most effectively saturates the GPU cores. 

\paragraph{Batched Execution.} Upon selecting the target type $\tau^*$, the execution engine performs a \textit{Cross-Query Operator Fusion}. Let $\{o_1, o_2, \dots, o_k\}$ be the set of operators in $\mathcal{R}_t$ with $\text{type}(o_i) = \tau^*$. Regardless of their original query contexts, their input tensors $\mathbf{X} = \{\mathbf{x}_1, \mathbf{x}_2, \dots, \mathbf{x}_k\}$ are coalesced into a contiguous memory block to form a unified batch $\mathbf{X}_{batch} \in \mathbb{R}^{k \times d}$. 
This unified representation allows the engine to invoke a single high-performance GPU kernel:
\begin{equation}
    \mathbf{Y}_{batch} = \text{Kernel}_{\tau^*}(\mathbf{X}_{batch}; \theta_{\tau^*})
\end{equation}
where $\theta_{\tau^*}$ represents the shared parameters for the operator type. This mechanism effectively mitigates kernel launch overhead and eliminates branch divergence across heterogeneous query structures, ensuring that the hardware operates near its theoretical peak performance.

\begin{figure}[t]
    \centering
    % \vspace{-3mm}
    \includegraphics[width=1.0\linewidth]{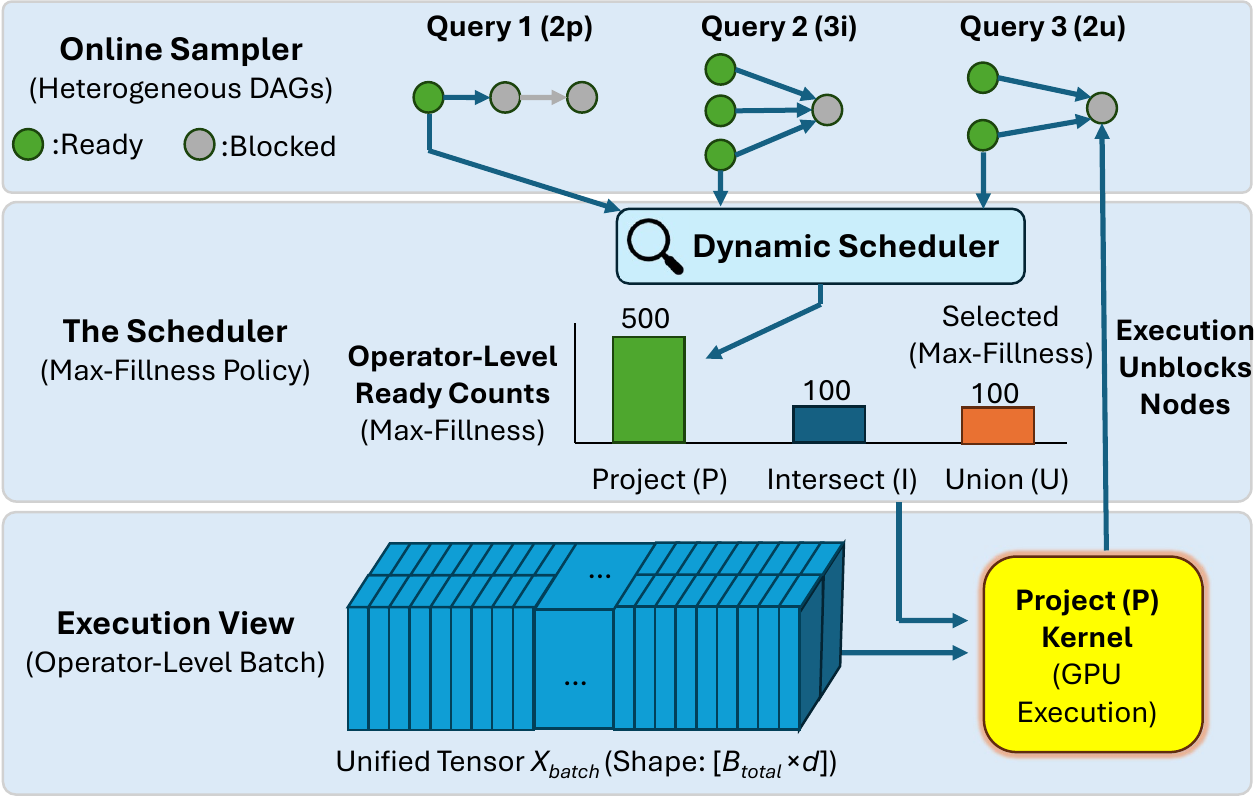}
% \caption{
%     \textbf{The illustration of the dynamic scheduling of \method{} in Secion \ref{sec:dynamic_scheduling}.} 
%     Workflow of the Max-Fillness Dynamic Scheduler. The scheduler orchestrates the data flow by monitoring the "ready state" of operators across heterogeneous query DAGs. (Mechanism): At each step, it prioritizes the operator type with the highest workload accumulation (e.g., Project pool has 500 tasks) to saturate GPU cores. (Outcome): This dynamic selection effectively unblocks dependent nodes, transforming the training process into a continuous, high-throughput stream.
    
% }
\caption{
    \textbf{Max-Fillness Dynamic Scheduling.} 
    By prioritizing operator pools with the highest workload, the scheduler maximizes GPU utils. This dynamic selection effectively unblocks dependent nodes, ensuring a continuous, high-throughput training stream.
}
    \vspace{-3mm}
    \label{fig:dynamic_scheduling}
\end{figure}

\subsection{Vectorized Objective Optimization}

\label{sec:vectorized_objective_optimization}

% \begin{figure}[h]
%     \centering
%     % \vspace{-3mm}
%     \includegraphics[width=1.0\linewidth]{fig/vectorized_objective_optimization.pdf}
% \caption{
%     \textbf{The illustration of vectorized objective optimization of \method{} in Secion \ref{sec:vectorized_objective_optimization}.} 
% }
%     % \vspace{-4mm}
%     \label{fig:vectorized_objective_optimization}
% \end{figure}

Building on the QueryDAG abstraction, we implement a multi-phase optimization strategy that restructures the entire training lifecycle to maximize system throughput.

\paragraph{Vectorized Logit Formulation.} 
Beyond structural scheduling, the computation of the objective function often suffers from high latency due to iterative embedding lookups. We reformulate the loss computation to leverage high-bandwidth memory (HBM). Instead of computing the scores for positive and negative samples individually, we represent the query embeddings and candidate entity embeddings as tensors $\mathbf{Q} \in \mathbb{R}^{B \times d}$ and $\mathbf{E} \in \mathbb{R}^{N \times d}$, respectively. The objective function calculation is then cast as a dense matrix-vector product:
\begin{equation}
    \mathcal{L} = \sum_{i=1}^{B} \psi(\mathbf{Q}_i \cdot \mathbf{E}_{pos}^\top) + \sum_{i=1}^{B} \sum_{j \in \mathcal{N}_i} \psi(\mathbf{Q}_i \cdot \mathbf{E}_{neg, j}^\top)
\end{equation}
where $\psi(\cdot)$ denotes the scoring function (e.g., Distance or Log-Sigmoid). This formulation allows the underlying linear algebra libraries to optimize data reuse via shared memory.

\paragraph{Efficient Tensor Indexing.} 
To further alleviate CPU-GPU synchronization bottlenecks, we introduce a \textit{Precomputed Indexing} mechanism. By pre-calculating the memory offsets for entity embeddings within the QueryDAG, we employ asynchronous scatter/gather operations. This replaces redundant tensor concatenation and splitting, ensuring that the critical path of the forward and backward passes remains entirely within the accelerator's memory space.

\subsection{Efficient System Architecture and Optimization}
\label{sec:method_optimization}

To facilitate training on large-scale data with high-throughput requirements, we propose a multi-faceted system optimization framework. This framework addresses two primary bottlenecks: memory-intensive intermediate tensor management and sub-optimal hardware utilization during complex query reasoning.

% \paragraph{Proactive Memory Management via Reference Counting.}
% \label{sec:memory_mgmt}
% To enable training with massive batch sizes, we implement Eager Reference Counting to achieve fine-grained resource reclamation. Formally, for any intermediate tensor $T$ generated within a Query Directed Acyclic Graph (QueryDAG), let $\mathcal{V}_{\text{desc}}(T)$ denote the set of descendant operators that consume $T$ as input. The system maintains a dynamic reference counter. A tensor $T$ is eligible for immediate memory deallocation if and only if all dependent operations have been executed:
% \begin{equation}
% \text{Reclaim}(T) \iff \sum_{v \in \mathcal{V}_{op}} \mathbb{I}(T \in \text{input}(v)) = 0
% \end{equation}
% By strictly enforcing this reclamation policy, our system significantly lowers the memory pressure compared to query-scoped allocation, enabling larger batch sizes training without triggering out-of-memory errors on modern hardware. 

\paragraph{Proactive Memory Management via Reference Counting.}
\label{sec:memory_mgmt}
To enable training with massive batch sizes, we implement
Eager Reference Counting to achieve fine-grained resource
reclamation. Formally, for any intermediate tensor $T$
generated within a Query Directed Acyclic Graph (QueryDAG),
let $\mathcal{V}_{\text{desc}}(T)$ denote the set of downstream operators
that consume $T$ as input, and let $\mathcal{F}_t \subseteq \mathcal{V}$
denote the set of operators that have been executed by
scheduling step~$t$. The system maintains a dynamic
reference counter for each tensor. A tensor $T$ is eligible
for immediate memory deallocation at step $t$ if and only
if every consumer has already been executed:
\begin{equation}
  \textsc{Reclaim}(T) \iff
  \sum_{v \,\in\, \mathcal{V}_{\text{desc}}(T)}
    \mathbb{I}\!\bigl(v \notin \mathcal{F}_t\bigr) = 0
\end{equation}
By strictly enforcing this reclamation policy, our system significantly lowers the memory pressure compared to query-scoped allocation, enabling larger batch sizes training without triggering out-of-memory errors on modern hardware. 

\paragraph{Asynchronous Parallelism and Pipelining.}
\label{sec:parallelism}
The memory efficiency gains from reference counting serve as a foundation for high-concurrency execution. We implement a decoupled execution pipeline characterized by two key parallelization strategies:

 \textit{Multi-stream Operator Execution:} We leverage independent CUDA streams to parallelize disjoint operator types within the QueryDAG. This allows for the simultaneous execution of compute-bound and memory-bound kernels, effectively masking hardware latency.
 \textit{Heterogeneous Pipelining:} To handle massive graphs exceeding GPU capacity, we adopt an asynchronous CPU-GPU pipelining strategy inspired by SMORE~\citep{ren2022smore}. Feature and structure embeddings are stored in host memory (CPU) and asynchronously fetched to pinned memory before scatter-gathering to the GPU. To mitigate the bottleneck of CPU-side data preparation, we use a consumer-producer pipeline. While the GPU executes current operator batches, the CPU concurrently performs asynchronous sampling for subsequent queries, ensuring a continuous data flow to the accelerators. 

% \paragraph{Asynchronous Parallelism and Pipelining.}
% \label{sec:parallelism}
% We implement a decoupled execution pipeline characterized by two key parallelization strategies: Multi-stream Operator Execution and Heterogeneous CPU-GPU Pipelining. More details are in Appendix Section~\ref{app:Implementation_Details}.

\paragraph{Efficiency-oriented Intersect and Union Operator.} 
\label{sec:efficient_inter_union}

\begin{figure}[t]
    \centering
    % \vspace{-3mm}
    \includegraphics[width=1.0\linewidth]{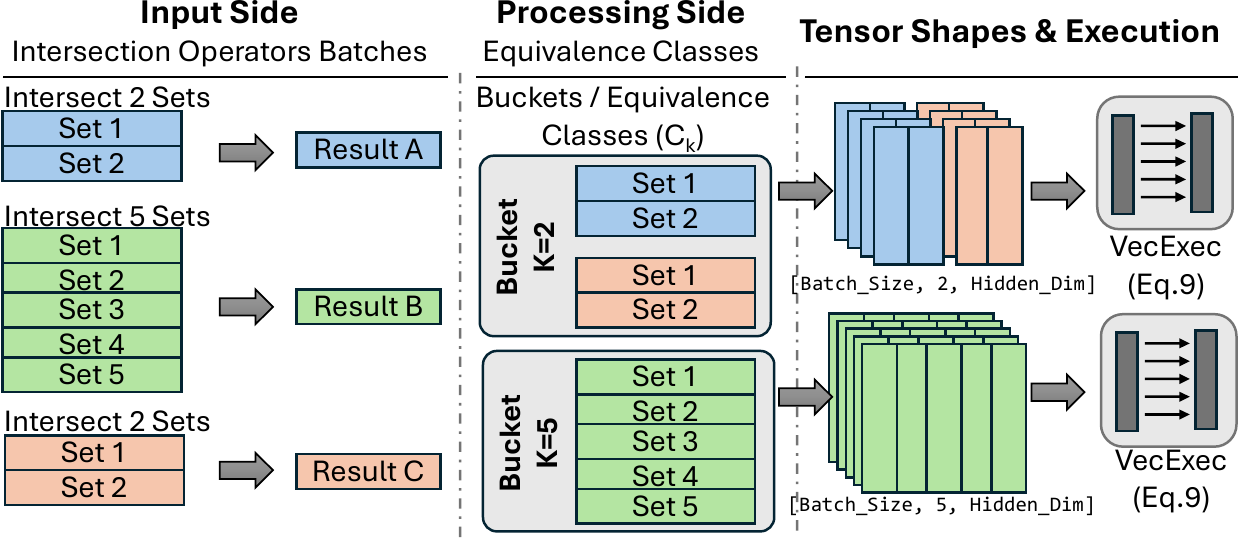}
\caption{
    \textbf{Vectorized execution for Intersection/Union.} 
    By grouping operations into cardinality-based Equivalence Classes, we eliminate tensor misalignment.
    This enables perfectly aligned \textbf{Vectorized Execution}, replacing slow loops with dense matrix operations.
}
    \vspace{-5mm}
    \label{fig:efficient_inter_union}
\end{figure}

As shown in Figure \ref{fig:efficient_inter_union}, unlike structural operators with fixed input degrees (e.g., \textit{Projection} or \textit{Negation}), the \textit{Intersection} ($\cap$) and \textit{Union} ($\cup$) operators in logical reasoning encounter variable-cardinality inputs, which typically leads to fragmented memory allocation and sub-optimal GPU utilization. To handle the computational complexity of the Intersect and Union Operator, we propose a batching strategy based on input cardinality. Let $\Phi \in \{\mathcal{O}_{int}, \mathcal{O}_{un}\}$ denote a set-based operator acting on an arbitrary number of input embeddings $\mathcal{S} = \{\mathbf{e}_1, \dots, \mathbf{e}_n\}$. We define the computation over a query batch as a partition into equivalence classes $\{\mathcal{C}_k\}_{k \in \mathcal{K}}$, where each class $\mathcal{C}_k$ clusters operations with identical input cardinality $k = |\mathcal{S}|$:
\begin{equation}
    \mathcal{C}_k = \{ \phi_i \in \text{Batch} \mid \text{card}(\text{input}(\phi_i)) = k \}.
\end{equation}
The global operation is then reformulated as a sequence of vectorized executions:
\begin{equation}
    \mathcal{O}_{\Phi} = \bigoplus_{k \in \mathcal{K}} \text{VecExec}(\mathcal{C}_k),
\end{equation}
where $\text{VecExec}(\cdot)$ denotes a synchronized tensor operation on a $k$-depth stacked tensor. This grouping mechanism optimizes GPU memory utilization and maximizes throughput by reducing fragmented memory allocations.

\paragraph{Online Data Sampling and Test Time Optimization.}
For NGDBs on large-scale graphs, the query space is effectively infinite, making the storage of pre-computed datasets prohibitive. Consequently, we employ an online data sampling system that generates queries dynamically during training. Beyond eliminating storage overhead, this approach exploits temporal locality in query streams. By adjusting the sampling distribution on-the-fly based on query difficulty, the system can optimize for specific query patterns that appear in bursts during multi-run training sessions. This dynamic generation allows for an unlimited stream of diverse samples and enables the curriculum to adaptively prioritize complex multi-hop structures to stabilize convergence.

\subsection{Efficient Integration of Semantic Entity Representations}
\label{sec:semantic_embedding}

\begin{figure}[t]
    \centering
    % \vspace{-3mm}
    \includegraphics[width=1.0\linewidth]{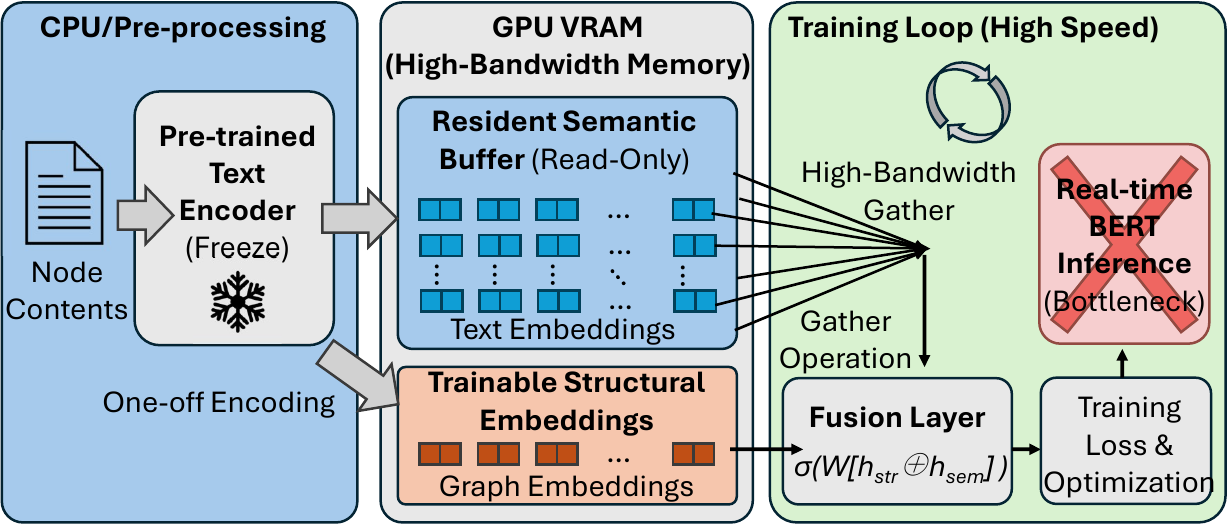}
\caption{\textbf{Illustration of the decoupled semantic integration strategy.} 
        By relegating the heavy text encoder (PTE) to an offline pre-computation phase and caching embeddings in GPU memory, \method{} achieves minimal-overhead semantic fusion during training.
        } 
    \vspace{-3mm}
    \label{fig:semantic_embedding}
\end{figure}

To augment entities with rich textual knowledge, we integrate high-dimensional semantic priors derived from Pre-trained Text Encoders (PTEs). However, joint training with PTEs typically incurs prohibitive overheads, transforming high-throughput graph reasoning into a compute-bound or I/O-bound bottleneck. To resolve this, we propose a \textit{decoupled, GPU-resident integration strategy} that ensures the training loop remains strictly inference-free.

\paragraph{Decoupled Semantic Encoding.} 
Instead of performing real-time feature extraction, which couples the heavy PTE inference with the gradient descent loop, we relegate semantic acquisition to an \textit{offline pre-computation phase}. Given an entity set $\mathcal{E}$, we employ a frozen PTE $\Phi(\cdot)$ to map each entity $e_i \in \mathcal{E}$ to a high-dimensional latent space:
\begin{equation}
    \mathcal{H}^{sem} = \{ \Phi(T_i) \mid e_i \in \mathcal{E} \} \in \mathbb{R}^{|\mathcal{E}| \times d_l}
\end{equation}
where $T_i$ denotes the textual description of entity $e_i$. Crucially, once $\mathcal{H}^{sem}$ is generated, the PTE $\Phi(\cdot)$ is \textbf{unloaded} from memory. This decoupling ensures that the subsequent training phase is completely agnostic to the complexity of the text encoder.

\paragraph{High-Throughput GPU-Resident Integration.} 
To eliminate the latency of CPU-to-GPU data transfer (PCIe bandwidth bottleneck), we implement a \textit{GPU-resident caching mechanism}. The global semantic manifold $\mathcal{H}^{sem}$ is registered as a persistent, non-trainable buffer directly within the GPU's High Bandwidth Memory (HBM).

% For a given mini-batch of entity indices $\mathcal{I} \subseteq \{1, \dots, |\mathcal{E}|\}$, the integration of semantic features is reformulated as a high-speed tensor indexing operation:
% \begin{equation}
%     \mathbf{E}_{batch}^{sem} = \text{Gather}(\mathcal{H}^{sem}, \mathcal{I})
% \end{equation}
% By maintaining the entire embedding table on-device, we circumvent the latency associated with frequent CPU-to-GPU data transfers and PCIe bandwidth limitations. To ensure the compatibility of these high-dimensional priors with the model's internal representations, we apply a linear projection layer $\mathcal{F}: \mathbb{R}^{d_l} \to \mathbb{R}^d$:
% \begin{equation}
%     \mathbf{e}_i^{fused} = \sigma \left( \mathbf{W}_p [\mathbf{h}_i^{str} \oplus \mathcal{F}(\mathbf{h}_i^{sem})] + \mathbf{b}_p \right)
% \end{equation}
% where $\mathbf{h}_i^{str}$ represents the structural embedding, $\oplus$ denotes the concatenation operator, and $\sigma$ is a non-linear activation function. This architecture ensures that the structural and semantic information are aligned within a unified computational space while maintaining near-optimal training efficiency.

Consequently, for a mini-batch of entity indices $\mathcal{I}$, the integration of semantic features is reduced from a model inference task to a high-speed tensor indexing operation:
\begin{equation}
    \mathbf{E}_{batch}^{sem} = \text{Gather}(\mathcal{H}^{sem}, \mathcal{I})
\end{equation}
This design achieves \textit{minimal-overhead} data access. To align these high-dimensional priors with the structural embedding space, we employ a lightweight linear projection $\mathcal{F}: \mathbb{R}^{d_l} \to \mathbb{R}^d$:
\begin{equation}
    \mathbf{e}_i^{fused} = \sigma \left( \mathbf{W}_p [\mathbf{h}_i^{str} \oplus \mathcal{F}(\mathbf{h}_i^{sem})] + \mathbf{b}_p \right)
\end{equation}
where $\oplus$ denotes concatenation and $\mathbf{h}_i^{str}$ is the learnable structural embedding. By treating semantic integration as a pure memory lookup, our framework sustains peak GPU utilization regardless of the PTE's original size.

\section{Experiments}

\subsection{Experimental Setup}
\paragraph{Baselines training framework.} 
We compare our single-hop KG completion runtime with state-of-the-art frameworks SQE~\citep{bai2023sequential}, SMORE~\citep{ren2022smore},  DGL-KE~\citep{zheng2020dgl}, Pytorch-Biggraph (PBG)~\citep{lerer2019pytorch}, and Marius~\citep{mohoney2021marius}. Then we evaluate the end-to-end training performance of \method{} on multi-hop reasoning over both small and large KGs. 

\paragraph{Backbone models.}
To validate the effectiveness of our training framework across diverse representational paradigms, we benchmark against a comprehensive suite of query encoding models. These baselines parameterize logical operators using various geometric and neural structures:
\textbf{GQE}~\citep{hamilton2018embedding} and \textbf{Q2P}~\citep{bai2022query2particles} utilize vector-based representations;
\textbf{Q2B}~\citep{ren2020query2box} and 
\textbf{BetaE}~\citep{ren2020beta} leverage Beta distributions for probabilistic reasoning; and
\textbf{FuzzQE}~\citep{chen2022fuzzy} applies fuzzy logic principles. For pretrained text encoder, we use Qwen3-Embedding-0.6B~\citep{qwen3embedding} and BGE-Base-En-v1.5~\citep{bge_embedding}.

\paragraph{Dataset.} 
Experiments were conducted on following benchmarks: FB15k, FB15k-237, NELL \citep{ren2020beta}, FB400k~\cite{talmor2018web}, ogbl-wikikg2~\cite{hu2020open}, and Atlas-Wiki~\cite{bai2025autoschemakg}. We used the validation and test queries for the first three datasets. For ogbl-wikikg2, queries were randomly sampled using official edge splits. For FB400k, we utilized SPARQL annotations from CWQ as queries, following the data splitting method in SMORE~\citep{ren2022smore}. For ATLAS-Wiki-Triple-4M, queries were sampled using degree-weighted edge
sampling. Dataset statistics are listed in Table~\ref{tab:kg_stat} in Appendix~\ref{app:KG_statistics}. 

% \textbf{Training queries:} Generated online with weighted sampling. \textbf{Query types:} 14 types ($1p$, $2p$, $3p$, $2i$, $3i$, $pi$, $ip$, $2u$, $up$, $2in$, $3in$, $pin$, $pni$, $inp$).

% We used the three datasets FB15k, FB15k-237, NELL from \citet{ren2020beta}. These three KGs are small-scale with at most 60k entities. To create a set of large-scale multi-hop KG reasoning benchmarks, we further sample queries on three large KGs: FB400k, ogbl-wikikg2 and Freebase. 
% Ogbl-wikikg2 is a KG from the Open Graph Benchmark~\cite{hu2020open}. 
% FB400k is a subset of Freebase~\cite{bollacker2008freebase} which is derived based on a knowledge graph question answering dataset ComplexWebQuestion (CWQ)~\cite{talmor2018web}. 
% We further look at the complete Freebase KG used in DGL-KE. 

\paragraph{Framework and Hardware.}
Our framework is implemented using PyTorch 2.0 and Python 3.10. Code will be released upon publication. Unless otherwise stated, all other experiments are conducted on NVIDIA A6000 GPUs (48GB memory) with 40 CPU cores.

\subsection{Scalability}
\paragraph{GPU memory and end-to-end training speed.} We test \method{} on all six KGs and report the end-to-end training speed and GPU memory data. 
As reported in Table~\ref{tab:calibrate}, \method{} achieves substantial throughput gains compared to representative baselines. For instance, on the FB15k dataset using the BetaE model, our system attains 4,477 queries/sec, representing a 7.0$\times$ speedup over the SQE framework.
While \method{} exhibits higher GPU memory footprints compared to SMORE due to our GPU-resident caching strategy, the memory usage remains well within the capacity of modern accelerators and often stays significantly lower than SQE in complex scenarios like NELL995. As shown in Table~\ref{tab:large_kg_comprehensive}, this efficiency extends to massive KGs, where \method{} sustains high throughput even with million-level entities, proving its viability for production-scale neural graph databases.

% \begin{table}[h]
% \centering
% \caption{Performance on three large KGs.\label{tab:large_KG_GPU}}
% \setlength{\tabcolsep}{3pt}
% \resizebox{0.48\textwidth}{!}{%
% % 默认是 6pt，缩小到 3pt 可以显著减宽
% \begin{tabular}{llcc} % 修正为实际的 4 列
% \toprule
% Model & Dataset & Queries (x1K/sec) & GPU Mem (GB) \\
% \midrule
% \multirow{3}{*}{BetaE} 
%  & FB400k & 19.97 & 14 \\
%  & ogbl-wikikg2 & 19.65 & 14 \\
%  & Atlas-Wiki-Triple-4M & 17.89 & 15\\
% \midrule
% \multirow{3}{*}{Q2B} 
%  & FB400k & 21.55 & 11 \\
%  & ogbl-wikikg2 & 20.75 & 11 \\
%  & Atlas-Wiki-Triple-4M & 18.47 & 12 \\
% \midrule
% \multirow{3}{*}{GQE} 
%  & FB400k & 24.68 & 7.5 \\
%  & ogbl-wikikg2 & 23.92 & 8 \\
%  & Atlas-Wiki-Triple-4M & 22.00 & 10 \\
% \bottomrule
% \end{tabular}
% }
% \vspace{-4mm}
% \end{table}

\begin{table}[t] 
\centering
\caption{\textbf{Scalability and Predictive Performance on Massive Knowledge Graphs.} We report the filtered MRR (\%), training throughput ($10^3$ Queries/sec), and peak GPU memory occupancy (GB) across three production-scale datasets. \method{} enables high throughput and robust reasoning across diverse paradigms on production-scale graphs with millions of entities.
}
\label{tab:large_kg_comprehensive}
\small
\setlength{\tabcolsep}{5pt} % 适度调节间距
\begin{tabular}{llccc}
\toprule
\rowcolor{lightgray}
\textbf{Dataset} & \textbf{Model} & \textbf{MRR (\%)} & \textbf{TPut ($\uparrow$)} & \textbf{Mem (GB $\downarrow$)} \\
\midrule
\multirow{3}{*}{FB400k} 
  & GQE   & 35.84 & \textbf{24.68} & \textbf{7.5} \\
  & Q2B   & \textbf{52.33} & 21.55 & 11.0 \\
  & BetaE & 50.40 & 19.97 & 14.0 \\
\midrule
\multirow{3}{*}{ogbl-wikikg2} 
  & GQE   & 32.88 & \textbf{23.92} & \textbf{8.0} \\
  & Q2B   & 42.01 & 20.75 & 11.0 \\
  & BetaE & \textbf{44.54} & 19.65 & 14.0 \\
\midrule
\multirow{3}{*}{\shortstack[l]{ATLAS-Wiki\\-Triple-4M}} 
  & GQE   & 7.31  & \textbf{22.00} & \textbf{10.0} \\
  & Q2B   & \textbf{9.22}  & 18.47 & 12.0 \\
  & BetaE & 9.01  & 17.89 & 15.0 \\
\bottomrule
\end{tabular}
\end{table}

% \paragraph{GPU utilization.} In Figure~\ref{fig:gpu_util} we further show the GPU utilization of BetaE model with different multi-hop query structures on Freebase. We plot the average utilization of 8 GPUs with smoothing of 10 seconds. More complex queries like \texttt{3p} and \texttt{pni} would have higher fluctuation due to the variance of sampling time. However as these queries also require more neural ops which in turn brings up the GPU utilization. Generally our system is able to keep a high GPU utilization for all these query structures.

\begin{figure}[t] % [t] 表示优先放在页首，单栏图通常放页首或页尾
    \centering
    % width=\linewidth 确保图片撑满单栏的宽度
    \includegraphics[width=\linewidth]{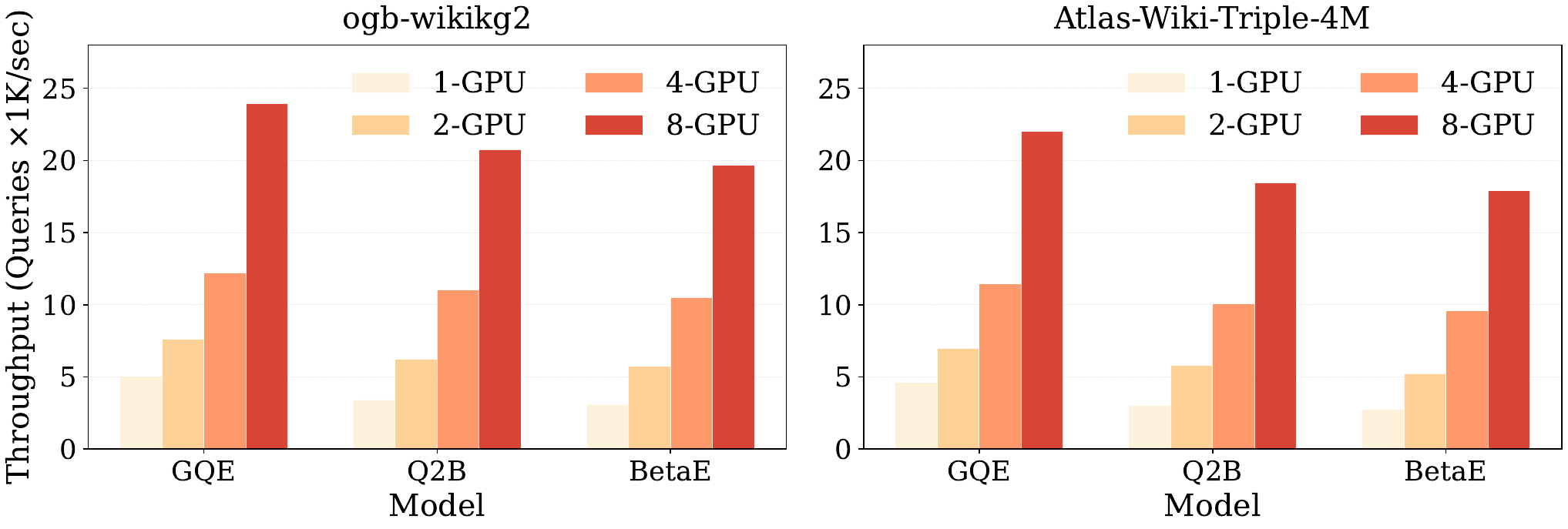}
    \caption{\textbf{Multi-GPU Throughput Scalability.} 
    Training throughput increases near-linearly with the number of GPUs across ogbl-wikikg2 and ATLAS-Wiki-Triple-4M. 
    Units are in \# Queries (x1K/sec).}
    \label{fig:gpu_scaling}
        \vspace{-4mm}
\end{figure}

\paragraph{Multi-GPU speed up.}
We report the training speed-up on 1, 2, 4, and 8 GPUs across different methods on the ogbl-wikikg2 and ATLAS-Wiki-Triple-4M in Figure~\ref{fig:gpu_scaling}. The training throughput increases almost linearly with the number of GPUs, demonstrating that our toolkit is highly scalable and the communication overhead is minimal.

\paragraph{Single-hop link prediction completion runtime.} Here we compare the single-hop link prediction (KG completion) runtime performance with state-of-the-art large-scale KG frameworks including Marius, PBG and SMORE. We report the results for ComplEx model with 100 embedding dimension on Freebase. We use the same multi-GPU V100 configuration as in Marius, and the current official release versions are also the same. For baseline results, we reuse Table 5 from~\citet{ren2022smore}. As shown in Table~\ref{tab:link_prediction}, \method{} achieves significantly faster runtime in 1-GPU setting than PBG. Also it scales better than the other systems, while Marius does not officially support multi-GPU parallel training functionality at the current stage. 

% Given that \method{} adopts the design choice of synchronized gradient update for dense parameters, and we do not partition the graph for the sake of multi-hop reasoning, it is nontrivial to be still comparable in the single-hop case.

\begin{table}[t]
\centering
\caption{Runtime performance on Freebase KG. *Results taken from ~\citet{mohoney2021marius} and ~\citet{ren2022smore} with the same V100 GPUs. \label{tab:link_prediction}
}
\vspace{-3mm}
\resizebox{0.48\textwidth}{!}{%
    \begin{tabular}{c|c|ccc}
    \toprule
    \multirow{2}{*}{System} &  \multicolumn{4}{c}{Epoch Time (s)} \\
    & 1-GPU & 2-GPU & 4-GPU & 8-GPU  \\
    \hline
    Marius~\citep{mohoney2021marius}* & 727 &  - & - & - \\
    PBG~\citep{lerer2019pytorch}* & 3060 & 1400 & 515 & 419 \\
    SMORE~\citep{ren2022smore}* & 760 & 411 & 224 & 121 \\
    \method{} (Ours) & \textbf{628} & \textbf{322} & \textbf{181} & \textbf{94} \\
    \bottomrule
    \end{tabular}
}
\end{table}

\subsection{Predictive performance}
\paragraph{Calibration on small KGs.} We first evaluate our \method{} system for five representational paradigms (BetaE, Q2B, GQE, Q2P, and FuzzQE) on the three small benchmark datasets created in BetaE. For a fair comparison, we test our system against other frameworks under identical online sampler.
To determine the system's throughput, we record both the data loading and sampling durations during this process. 
As summarized in Table~\ref{tab:calibrate}, \method{} consistently outperforms or matches competitive baselines in terms of MRR, demonstrating that training on dynamic online streams does not compromise reasoning accuracy. 

\begin{table*}[t]
\centering
\caption{Performance comparison of \method{} vs. KGReasoning, SMORE and SQE.}
\label{tab:calibrate}
\resizebox{\textwidth}{!}{%
\renewcommand{\arraystretch}{0.85} % 保持紧凑的行高
% 列定义：l l | cccc | cccc | cccc (共14列)
\begin{tabular}{l l | cccc | cccc | cccc}
\toprule
% 表头第一行：背景灰，指标名称跨4列
\rowcolor{lightgray}
& & \multicolumn{4}{c}{MRR (\%)} & \multicolumn{4}{c}{Training queries (Queries/Sec)} & \multicolumn{4}{c}{GPU Memory (GB)} \\
\cline{3-6} \cline{7-10} \cline{11-14}
\rowcolor{lightgray}
\multirow{-2}{*}{Dataset} & \multirow{-2}{*}{Model} & KGR & SMORE & SQE & \method{} & KGR & SMORE & SQE & \method{} & KGR & SMORE & SQE & \method{} \\
\midrule

% Data Block 1: FB15k (注意：这里由3改为5，因为有5个模型)
\multirow{5}{*}{FB15k} 
  & BetaE & 41.60 & 40.39 & 40.82 & \textbf{43.04} & 291 & 2808 & 636 & \textbf{4477} & 2.68 & \textbf{1.16} & 8.05 & 12.95 \\
  & Q2B   & 38.00 & 41.54 & 41.08 & \textbf{42.85} & 855 & 3588 & 343 & \textbf{4086} & 1.32 & \textbf{0.96} & 6.01 & 5.07 \\
  & GQE   & 28.00 & \textbf{30.60} & 29.21 & 30.39 & 1395 & 3770 & 4598 & \textbf{6271} & \textbf{0.76} & 1.09 & 8.03 & 11.33 \\
  & Q2P   & - & - & 44.30 & \textbf{46.90} & - & - & 832 & \textbf{1940} & - & - & \textbf{4.76} & 11.41 \\
  & FuzzQE   & - & - & 42.17 & \textbf{42.33} & - & - & 720 & \textbf{2973} & - & - & 20.14 & \textbf{8.45} \\

\midrule

% Data Block 2: FB15k-237
\multirow{5}{*}{FB15k-237} 
  & BetaE & 20.90 & 19.67 & 20.92 & \textbf{21.11} & 178 & 1633 & 655 & \textbf{4750} & 2.68 & \textbf{1.14} & 6.03 & 9.81 \\
  & Q2B   & 20.10 & \textbf{20.42} & 19.05 & 20.19 & 428 & 3115 & 343 & \textbf{4663} & 1.3 & \textbf{0.93} & 6.04 & 5.09 \\
  & GQE   & 16.30 & 15.68 & 16.75 & \textbf{16.98} & 1433 & 2882 & 1910 & \textbf{6034} & \textbf{0.75} & 1.02 & 8.01 & 9.77 \\
  & Q2P   & - & - & 20.12 & \textbf{22.08} & - & - & 842 & \textbf{1884} & - & - & \textbf{4.22} & 11.34 \\
  & FuzzQE   & - & - & 21.07 & \textbf{21.35} & - & - & 1350 & \textbf{2934} & - & - & 17.02 & \textbf{8.54} \\

\midrule

% Data Block 3: NELL995
\multirow{5}{*}{NELL995} 
  & BetaE & 24.60 & 23.17 & 23.43 & \textbf{24.90} & 166 & 1807 & 154 & \textbf{4640} & 3.42 & 1.28 & 35.02 & 6.81 \\
  & Q2B   & 22.90 & 21.84 & 22.21 & \textbf{23.79} & 765 & 1926 & 82 & \textbf{4521} & 1.5 & \textbf{1} & 25.04 & 5.14 \\
  & GQE   & 18.60 & 17.53 & 18.02 & \textbf{18.89} & 1224 & 3691 & 2959 & \textbf{6329} & \textbf{1.01} & 1.17 & 45.04 & 7.12 \\
  & Q2P   & - & - & 23.18 & \textbf{25.65} & - & - & 836 & \textbf{2309} & - & - & 9.69 & \textbf{9.56} \\
  & FuzzQE   & - & - & 24.31 & \textbf{25.88} & - & - & 2165 & \textbf{2680} & - & - & \textbf{8.07} & 8.16 \\

\midrule

% Average Row
\multicolumn{2}{l|}{Average Without Q2P/FuzzQE} & 25.67 & 25.65 & 25.72 & \textbf{26.90} & 748 & 2802 & 1298 & \textbf{5086} & 1.71 & \textbf{1.08} & 16.36 & 8.12 \\
\bottomrule
\end{tabular}%
}
\end{table*}

\paragraph{Query answering on large KGs.} We further evaluate \method{} on three large-scale benchmarks: FB400k, ogbl-wikikg2, and ATLAS-Wiki-4M. As summarized in Table~\ref{tab:large_kg_comprehensive}, \method{} successfully scales diverse representational paradigms to graphs with millions of entities. On ogbl-wikikg2, BetaE attains the highest performance with an MRR of 44.54\%, while Q2B demonstrates superior expressivity on FB400k, achieving 52.33\%. This shows our system maintains robust reasoning even on the massive graph which contains over 4 million entities.

% \begin{figure*}[h!]
% \centering
% % 左侧 minipage：放入原本的 fig/2.pdf
% \begin{minipage}{0.50\textwidth}
%     \centering
%     \includegraphics[width=0.99\textwidth]{fig/gpu_scaling_combined.pdf}
%     \vspace{-2mm}
%     \caption{\textbf{Multi-GPU Throughput Scalability.}
%     Training throughput increases near-linearly with the number of GPUs across ogbl-wikikg2 and ATLAS-Wiki-Triple-4M.
%     % Throughput comparison on Ogbl-wikikg2 and Atlas-Wiki-Triple-4M dataset across different GPU configurations. 
%     Units are in \# Queries (x1K/sec). \label{fig:gpu_scaling}}
% \end{minipage}
% \hfill
% \begin{minipage}{0.46\textwidth}
%     \centering
%     % width=\linewidth 确保图片充满这个 minipage 的宽度
%     \includegraphics[width=\linewidth]{fig/adaptive_sampling_grouped_by_dataset.pdf}
%     \caption{Performance comparison with and without Adaptive Sampling. Adaptive Sampling consistently improve MRR over all models and datasets. \label{fig:Adaptive_Sampling}}
% \end{minipage}

% % 右侧 minipage：保持原有的 gpu_scaling_combined.pdf
% \end{figure*}

\begin{figure*}[h]
\centering
\includegraphics[width=1.0\textwidth]{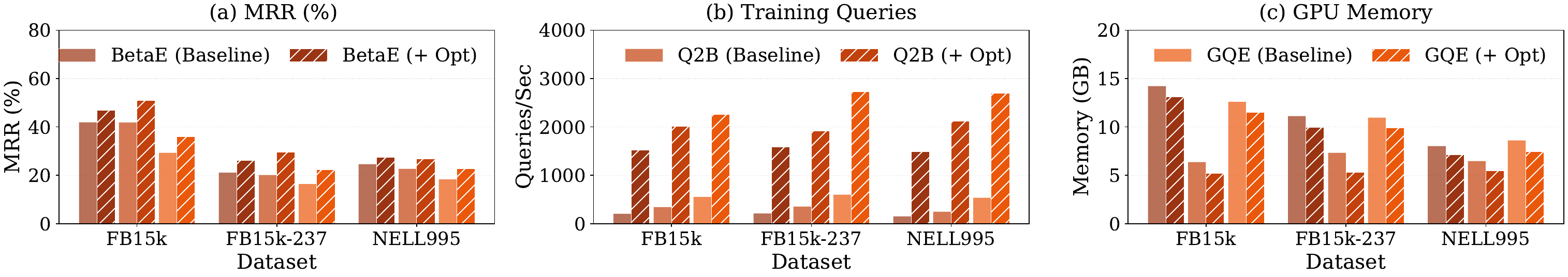}
\caption{\textbf{Impact of Decoupled Semantic Integration.} 
We compare the standard joint training (Solid) against our GPU-Resident Decoupled strategy (Hatched //) across three datasets and models. 
\textbf{(a) Predictive Performance:} Our method consistently improves MRR than without semantic priors by leveraging semantic priors. 
\textbf{(b) Throughput:} By removing the heavy Text Encoder from the training loop, we achieve a \textbf{5x-7x speedup}. 
\textbf{(c) Memory:} Despite caching embeddings, peak memory is reduced because the encoder is unloaded. 
Note: Higher is better for (a)\&(b); lower is better for (c).}
\label{fig:qwen}
    \vspace{-2mm}
\end{figure*}

\begin{figure}[t]
    \centering
    \includegraphics[width=\linewidth]{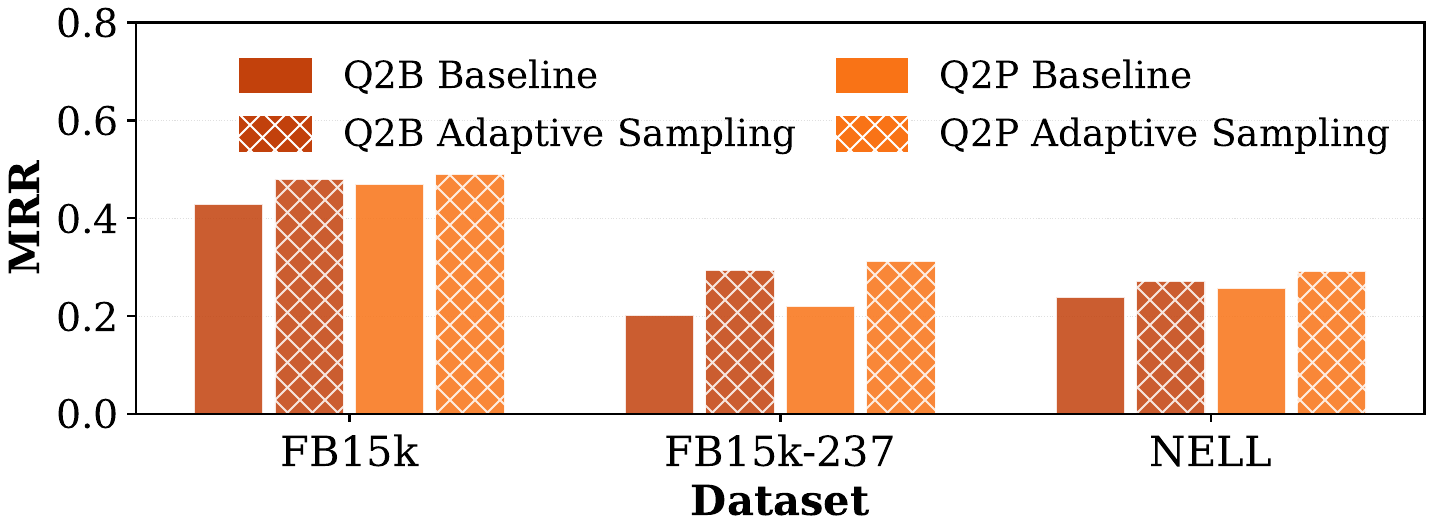}
    \caption{Performance comparison with and without Adaptive Sampling. 
    Adaptive Sampling consistently improves MRR over all models and datasets.}
    \label{fig:Adaptive_Sampling}
    \vspace{-3mm}
\end{figure}

% \begin{figure}[t]
% \centering
% \includegraphics[width=0.48\textwidth]{}
%     \caption{ GPU utilization with different query structures on Freebase with BetaE~\citep{ren2020beta}. \label{fig:gpu_util}  }
% \end{figure}

% \begin{figure}[t]
% \centering
% \includegraphics[width=0.48\textwidth]{fig/adaptive_sampling_grouped_by_dataset.pdf}
% \caption{Performance comparison with and without Adaptive Sampling. Adaptive Sampling consistently improve MRR over all models and datasets. \label{fig:Adaptive_Sampling} }
% \end{figure}

% \begin{table}[h!]
%     \centering
%     % Caption 放在表格上方是标准学术格式
%     \caption{\textbf{Scalability and Predictive Performance on Massive Knowledge Graphs.} \method{} enables robust reasoning across diverse paradigms on production-scale graphs with millions of entities.}
%     \label{tab:mergedmrr}
    
%     % 使用 \linewidth 让表格自动适应当前栏宽
%     % \resizebox{0.7\linewidth}{!}{%
%         \begin{tabular}{c|ccc}
%             \toprule
%             {\small Dataset \textbackslash\: Model} & GQE & Q2B & BetaE\\
%             \hline
%             FB400k & 35.84 & 52.33 & 50.40 \\
%             ogbl-wikikg2 & 32.88 & 42.01 & 44.54 \\
%             ATLAS-Wiki-4M & 7.31 & 9.22 & 9.01 \\
%             \bottomrule
%         \end{tabular}
%     % }%
% \end{table}

\paragraph{Online Sampling.}
\label{Online_Sampling}
To validate our adaptive approach, we conducted controlled experiments using a steered query distribution characterized by an abrupt spike in query difficulty every 15,000 training steps. As shown in Figure~\ref{fig:Adaptive_Sampling}, our adaptive sampling mechanism consistently outperforms the baseline across all configurations. It achieves an average relative improvement of 21.5\% in MRR. This indicates that adaptive sampling demonstrates superior robustness in handling non-stationary distributions by rapidly accommodating sudden difficulty shifts. This dynamic generation facilitates an unlimited stream of diverse samples, enabling the model to adaptively prioritize complex multi-hop structures and stabilize convergence in highly entropic environments.

% \begin{table}[t!]
%     \centering
%     \caption{Performance comparison with and without Adaptive Sampling. Adaptive Sampling consistently improves MRR over all models and datasets.}
%     \label{tab:online_sampling}
%     \setlength{\tabcolsep}{20pt} % 只有三列时，可以适当加大间距使表格不那么“缩”在中间
%     \resizebox{0.48\textwidth}{!}{%

%     \begin{tabular}{l l l}
%         \toprule
%         \textbf{Models} & \textbf{MRR (Base)} & \textbf{MRR (Adapt.)} \\
%         \midrule
%         \textit{Q2B-FB15k}     & 0.429 & 0.480 \\
%         \textit{Q2B-FB15k-237} & 0.202 & 0.293 \\
%         \textit{Q2B-NELL}      & 0.238 & 0.271 \\
%         \textit{Q2P-FB15k}     & 0.469 & 0.490 \\
%         \textit{Q2P-FB15k-237} & 0.221 & 0.312 \\
%         \textit{Q2P-NELL}      & 0.257 & 0.291 \\
%         \bottomrule
%     \end{tabular}
%     }
% \end{table}

\subsection{Impact of Decoupled Semantic Augmentation}

\paragraph{Throughput, Memory and Predictive Performance.} 
On standard benchmarks (FB15k, NELL) where the full manifold fits in VRAM, we utilize the GPU-Resident strategy to maximize throughput by eliminating PCIe latency.
We evaluate the impact of our decoupled semantic integration strategy across three models: BetaE, Q2B, and GQE. Figure~\ref{fig:qwen} illustrates the substantial performance boost observed with the Qwen3-Embedding. Furthermore, as detailed in Appendix Table~\ref{tab:with_qwen}, these trends remain consistent when utilizing BGE-Base-En-v1.5. Across both encoders, the optimizations consistently yield a 5x-7x throughput speedup and reduce memory consumption.
\textbf{Throughput:} By shifting the computationally expensive text encoder inference to an offline phase and caching embeddings in GPU memory, we achieve a \textbf{5$\times$--7$\times$ speedup} in training throughput. This confirms that our ``inference-free'' design effectively eliminates the computational bottleneck.
\textbf{Memory:} Despite caching the full entity manifold on-device, the overall GPU memory footprint decreases. This is because the massive Pre-trained Text Encoder (PTE) is completely \textbf{unloaded} from VRAM during training, offsetting the storage cost of the embedding buffer.
\textbf{Predictive Performance:} Semantic Entity Joint Training consistently improve MRR across all models and datasets by harnessing linguistic priors.

\section{Conclusion}
\label{sec:conclusion}

We presented \method{}, a framework resolving the efficiency-expressivity tension in Neural Graph Databases. By adopting operator-level batching, \method{} overcomes topological rigidity, achieving leading throughput and near-linear multi-GPU scalability. We also showed that integrating Pre-trained Text Encoders alleviates representation friction in sparse graphs. Despite a modest memory increase, \method{} sustains robust reasoning on massive graphs, offering a scalable foundation for next-generation neuro-symbolic systems.

% We presented \method{}, a high-throughput and semantically-augmented framework designed to resolve the fundamental tension between hardware efficiency and representational expressivity in Neural Graph Databases (NGDBs). By transitioning from query-level to operator-level batching, \method{} effectively mitigates topological rigidity, achieving a leading increase in throughput across diverse query workloads while maintaining near-linear scalability on multi-GPU configurations. Our investigation also demonstrates that external linguistic priors from Pre-trained Text Encoders can significantly alleviate representation friction in sparse knowledge structures. Furthermore, our methods provide critical insights for the community on scaling neuro-symbolic systems. While \method{} introduces a modest increase in memory footprint due to its caching strategy, its ability to sustain robust reasoning on massive graphs with over 4 million entities makes it a viable foundation for next-generation neural graph databases. 

\newpage
\clearpage

% Authors are kindly asked to make their submissions as accessible as possible
% for everyone including people with disabilities and sensory or neurological
% differences. Tips of how to achieve this and what to pay attention to will be
% provided on the conference website \url{http://icml.cc/}.

% \section*{Software and Data}

% If a paper is accepted, we strongly encourage the publication of software and
% data with the camera-ready version of the paper whenever appropriate. This can
% be done by including a URL in the camera-ready copy. However, \textbf{do not}
% include URLs that reveal your institution or identity in your submission for
% review. Instead, provide an anonymous URL or upload the material as
% ``Supplementary Material'' into the OpenReview reviewing system. Note that
% reviewers are not required to look at this material when writing their review.


\section*{Impact Statement}

This paper presents work whose goal is to advance the field of Machine
Learning. There are many potential societal consequences of our work, none
which we feel must be specifically highlighted here.


\bibliography{example_paper}
\bibliographystyle{icml2026}

%%%%%%%%%%%%%%%%%%%%%%%%%%%%%%%%%%%%%%%%%%%%%%%%%%%%%%%%%%%%%%%%%%%%%%%%%%%%%%%
%%%%%%%%%%%%%%%%%%%%%%%%%%%%%%%%%%%%%%%%%%%%%%%%%%%%%%%%%%%%%%%%%%%%%%%%%%%%%%%
% APPENDIX
%%%%%%%%%%%%%%%%%%%%%%%%%%%%%%%%%%%%%%%%%%%%%%%%%%%%%%%%%%%%%%%%%%%%%%%%%%%%%%%
%%%%%%%%%%%%%%%%%%%%%%%%%%%%%%%%%%%%%%%%%%%%%%%%%%%%%%%%%%%%%%%%%%%%%%%%%%%%%%%

\include{appendix}

\end{document}

%% file: appendix.tex
\newpage
\onecolumn
\appendix
\section{Limitations and Future Work}
Despite its efficiency, the current architecture relies on a frozen text encoder. Future research could explore lightweight end-to-end co-optimization of the semantic and structural subspaces. System Trade-offs. To scale to massive graphs with limited VRAM, our current implementation on 4M-node graphs relies on asynchronous CPU-offloading. While effective, this introduces a dependency on PCIe bandwidth, which may become a bottleneck for even larger graphs or larger batch sizes compared to a fully GPU-resident approach. Furthermore, the reliance on a frozen text encoder decouples semantic understanding from structural learning, potentially limiting the model's ability to adapt linguistic priors to specific graph topologies.
Additionally, extending the operator-level paradigm to streaming knowledge bases, where the topology evolves in real-time, remains a promising direction for achieving truly autonomous graph reasoning.

\subsection{Pretrained Embeddings for Structured Knowledge}

\noindent \textbf{Language Models for KGs.} Recent work has explored combining pretrained language models with knowledge graphs: KG-BERT~\citep{yao2019kg} uses BERT for triple classification. KEPLER~\citep{wang2021kepler} jointly trains language models with KG embeddings. SimKGC~\citep{wang2022simkgc} uses contrastive learning to align entity descriptions with KG structure. 
UltraQuery~\citep{galkin2024foundation} build pre-trained GNN foundation model for inductive reasoning on KGs.

\noindent \textbf{Semantic Embeddings.} Pretrained sentence transformers~\citep{wang2021tsdae} provide high-quality semantic embeddings. Recent models like Qwen3-Embedding~\citep{qwen3embedding} achieve state-of-the-art performance on semantic similarity tasks. However, integrating these semantic representations with complex multi-hop reasoning models remains an open challenge.

\label{sec:appendix}

\section{Algorithm}
In Algorithm~\ref{alg:operator_scheduling}, we presented our \method{} in the form of pseudo-code to illustrate how it works.

% \begin{algorithm}[h]
% \caption{Operator-Level Training Loop}
% \begin{algorithmic}[1]
% \Require Batch of Queries $\mathcal{Q}$, Model parameters $\Theta$
% \State Construct DAGs: $\mathcal{G} \leftarrow \{ \text{BuildDAG}(q) \mid q \in \mathcal{Q} \}$
% \State Initialize operator pools $\mathcal{P}_\tau$ for each type $\tau$
% \While{$\exists$ uncomputed nodes in $\mathcal{G}$}
%     \State $\mathcal{R} \leftarrow \{ v \in \mathcal{G} \mid \text{dependencies\_met}(v) \}$
%     \State Distribute $v \in \mathcal{R}$ into pools $\mathcal{P}_{\text{type}(v)}$
%     \State Select type $\tau^* \leftarrow \arg\max_{\tau} \rho(\tau)$ \Comment{Max-Fillness Policy}
%     \State Form batch $\mathcal{B} \leftarrow \text{PopBatch}(\mathcal{P}_{\tau^*}, B_{\max})$
%     \State $\mathbf{O} \leftarrow \text{ExecuteKernel}(\mathcal{B}, \Theta)$
%     \For{$o_i \in \mathcal{B}$}
%         \State Update node status: $o_i.\text{output} \leftarrow \mathbf{O}_i$
%         \State $\text{DecrementRefCounts}(\text{inputs}(o_i))$ \Comment{Eager Reclamation}
%     \EndFor
% \EndWhile
% \State Update $\Theta$ via backpropagation on root nodes
% \end{algorithmic}
% \label{alg:operator_scheduling}
% \end{algorithm}

\begin{algorithm}[h]
\caption{Operator-Level Dynamic Scheduling Training Loop}
\label{alg:operator_scheduling}
\begin{algorithmic}[1]
\Require Batch of Queries $\mathcal{Q}$, Model Parameters $\Theta$, Max Batch Size $B_{\max}$
\State $\mathcal{G} \leftarrow \bigcup_{q \in \mathcal{Q}} \textsc{BuildDAG}(q)$ \Comment{Construct fused computation graph}
\State $\mathcal{G} \leftarrow \mathcal{G} \cup \textsc{AddGradientNodes}(\mathcal{G})$ \Comment{Include backward operators}
\State Initialize operator pools $\mathcal{P}_\tau$ for each operator type $\tau$
\State Initialize ready set $\mathcal{R} \leftarrow \{ v \in \mathcal{G} \mid \text{indegree}(v) = 0 \}$

\While{$\mathcal{R}$ is not empty}
    \State Distribute $v \in \mathcal{R}$ into pools $\mathcal{P}_{\textsc{Type}(v)}$
    \State $\mathcal{R} \leftarrow \emptyset$ \Comment{Clear ready set, wait for next cycle}
    
    \State Select $\tau^* \leftarrow \arg\max_{\tau} \rho(\tau)$ \Comment{Apply \textbf{Max-Fillness Policy}}
    \State $\mathcal{B} \leftarrow \textsc{PopBatch}(\mathcal{P}_{\tau^*}, B_{\max})$
    
    \State $\mathbf{O} \leftarrow \textsc{ExecuteKernel}(\mathcal{B}, \Theta)$ \Comment{Fused GPU Kernel}
    
    \For{$o_i \in \mathcal{B}$}
        \State $o_i.\text{output} \leftarrow \mathbf{O}_i$
        \State $\textsc{DecrementRefCounts}(\textsc{Inputs}(o_i))$ \Comment{\textbf{Eager Reclamation}}
        \For{$v_{succ} \in \textsc{Successors}(o_i)$}
            \If{$\textsc{DependenciesMet}(v_{succ})$}
                \State $\mathcal{R} \leftarrow \mathcal{R} \cup \{v_{succ}\}$ \Comment{Update ready set}
            \EndIf
        \EndFor
    \EndFor
\EndWhile
\State \textsc{OptimizerStep}($\Theta$) \Comment{Apply gradients computed in the loop}
\end{algorithmic}
\end{algorithm}

\section{KG statistics}
In Table~\ref{tab:kg_stat}, we show the statistics of all six KGs we use with number of entities, relations, training, validation and test edges. For ATLAS-Wiki-Triple-Sample, we construct a 6.4\% subgraph from ATLAS-Wiki-Triple using degree-weighted edge sampling. Edge sampling probability is proportional to the sum of endpoint node degrees, prioritizing connections between high-degree (hub) nodes to preserve the graph's core topological structure.
\label{app:KG_statistics}
\begin{table*}[h]
\centering
\caption{KG statistics with number of entities, relations, training, validation and test edges.}
\label{tab:kg_stat}
% 使用 textwidth 占满双栏宽度，调整 arraystretch 增加行间距
\resizebox{\textwidth}{!}{%
\def\arraystretch{1.0}
\begin{tabular}{l r r r r r r}
\Xhline{1.2pt}
\rowcolor{tableheadcolor} % 表头背景色
\textbf{Dataset} & \textbf{Entities} & \textbf{Relations} & \textbf{Train Edges} & \textbf{Valid Edges} & \textbf{Test Edges} & \textbf{Total Edges}\\
\Xhline{1.2pt} % 既然参考用了 Xhline，这里保持一致
FB15k & 14,951 & 1,345 & 483,142 & 50,000 & 59,071 & 592,213\\
\rowcolor{gray!10} % 隔行变色
FB15k-237 & 14,505 & 237 & 272,115 & 17,526 & 20,438 & 310,079\\
NELL995 & 63,361 & 200 & 114,213 & 14,324 & 14,267 & 142,804\\
\rowcolor{gray!10} % 隔行变色
FB400k & 409,829 & 918 & 1,075,837 & 537,917 & 537,917 & 2,151,671 \\
ogbl-wikikg2 & 2,500,604 & 535 & 16,109,182 & 429,456 & 598,543 & 17,137,181 \\
\rowcolor{gray!10} % 隔行变色
% ATLAS-Wiki & 70,104,000 & - & 1,193,600,000 & 149,200,000 & 149,200,000 & 1,492,000,000 \\
ATLAS-Wiki-Triple-4M & 4,035,238 & 512,064 & 23,040,868  & 2,880,108 & 2,880,110 & 28,801,086  \\

% ATLAS-Wiki-Triple & 66,894,746 & - &  360,826,353 & 45,103,294 & 45,103,294 & 451,032,941  \\
% ATLAS-Wiki-Triple & 235,821,758 & 68,004,308 & 360,826,353 & 45,103,294 & 45,103,294 & 451,032,941 \\
% ATLAS-Wiki-Triple & 66,894,746 & 68,004,308 &  360,826,353 & 45,103,294 & 45,103,294 & 451,032,941  \\

% Freebase & 86,054,151 & 14,824 & 304,727,650 & 16,929,318 & 16,929,308 & 338,586,276 \\
\Xhline{1.2pt}
\end{tabular}%
}
\end{table*}

\section{Implementation Details}
\label{app:Implementation_Details}
\paragraph{Asynchronous Parallelism and Pipelining.}
\label{sec:parallelism}
We implement a decoupled execution pipeline characterized by two key parallelization strategies:

\begin{itemize}[leftmargin=*]
    \item \textbf{Multi-stream Operator Execution:} We leverage independent CUDA streams to parallelize disjoint operator types within the QueryDAG. This allows for the simultaneous execution of compute-bound and memory-bound kernels, effectively masking hardware latency.
    \item \textbf{Heterogeneous Pipelining:} To mitigate the bottleneck of CPU-side data preparation, we introduce a consumer-producer pipeline. While the GPU executes current operator batches, the CPU concurrently performs asynchronous sampling for subsequent queries, ensuring a continuous data flow to the accelerators.
\end{itemize}

% \paragraph{Parameter Constraint Projection.} 
% For probabilistic distributions within the model (e.g., the Beta distribution), we define a constrained parameter space $\mathcal{S}$ to prevent gradient instability and numerical overflow. Formally, for any distribution parameter $\theta \in \{\alpha, \beta\}$, we apply a projection function $\Pi(\cdot)$ such that:
% \begin{equation}
%     \hat{\theta} = \text{clip}(\theta, \epsilon, \tau)
% \end{equation}
% where $\epsilon = 0.05$ and $\tau = 10^9$ denote the lower and upper numerical bounds, respectively. This projection, coupled with entity-based regularization, ensures that the loss function remains bounded and prevents the occurrence of non-finite values (NaN) in the latent manifold.

\paragraph{Hyperparameters and Configuration.}Table~\ref{tab:hyperparams} summarizes the unified hyperparameter settings applied across all baseline query encoding models and the configuration for the Qwen3 semantic integration. To ensure a fair comparison with prior work, we maintain a consistent training environment: all baselines are initialized with a latent dimension of $400$ and trained using the same negative sampling strategy and optimizer configurations.

\begin{table}[t]
    \centering
    \caption{Hyperparameter settings and model configurations.}
    \label{tab:hyperparams}
    \begin{tabular}{l l}
        \toprule
        \textbf{Parameter} & \textbf{Value} \\
        \midrule
        Batch Size & $512$ queries \\
        Optimizer & Adam (lr $10^{-4}$) \\
        Margin ($\gamma$) & $12.0$ \\
        \bottomrule
    \end{tabular}
\end{table}

\section{Joint Structural-Semantic Representation}
\label{sec:joint_formulation}

We formalize the integration of semantic information not as a mere feature augmentation, but as a mapping into a \textbf{composite representation manifold} $\mathcal{H}_{\text{joint}}$. Specifically, we define the problem as finding a joint embedding for each entity $e_i \in \mathcal{E}$ that simultaneously inhabits two distinct latent subspaces.

\paragraph{Subspace Definitions.} 
The representation space is decomposed into:
\begin{enumerate}
    \item \textbf{Structural Subspace ($\mathcal{H}_{\text{str}} \in \mathbb{R}^{d_s}$):} An endogenous, learnable space optimized to satisfy the relational constraints and multi-hop logical dependencies inherent in $\mathcal{G}$.
    \item \textbf{Semantic Subspace ($\mathcal{H}_{\text{sem}} \in \mathbb{R}^{d_m}$):} An exogenous, fixed manifold derived from a pre-trained linguistic encoder $\Phi: \mathcal{D} \to \mathcal{H}_{\text{sem}}$. For each entity $e_i$, its semantic prior $\mathbf{z}_i = \Phi(d_i)$ is grounded in its textual description $d_i \in \mathcal{D}$.
\end{enumerate}

\paragraph{Problem Definition: Semantic Grounding.}
We posit that the complete representation of an entity $e_i$ is a function $\Psi$ that maps the Cartesian product of the two subspaces into the final embedding $\mathbf{x}_i$:
\begin{equation}
    \mathbf{x}_i = \Psi(\mathbf{h}_i, \mathbf{z}_i), \quad \text{s.t. } \mathbf{h}_i \in \mathcal{H}_{\text{str}}, \mathbf{z}_i \in \mathcal{H}_{\text{sem}}
\end{equation}
where $\mathbf{h}_i$ is a set of adaptive parameters and $\mathbf{z}_i$ acts as an invariant semantic anchor. 

\paragraph{Sufficient Statistics for Probabilistic Reasoning.}
For reasoning frameworks that model queries as probability distributions (e.g., box or beta-based distributions), the task is to define a density function $p(x | \theta_i)$. Under our joint formulation, the parameters $\theta_i$ (e.g., location and uncertainty measures) are defined as functionals of the joint space:
\begin{equation}
    \theta_i \propto \mathcal{F}(\mathbf{x}_i) = \mathcal{F}(\Psi(\mathbf{h}_i, \mathbf{z}_i))
\end{equation}
This formulation ensures that the resulting geometric shapes or distributions in the embedding space are intrinsically constrained by the linguistic grounding of the entities, even when the graph topology is sparse.

\section{Stochastic Online Query Sampling}
\label{app:online_sampler}

To handle the combinatorial explosion of query structures in Large-scale Knowledge Graphs (KGs), we implement an \textbf{Online Stochastic Sampler} $\mathcal{S}$. Unlike traditional offline methods that rely on pre-computed static datasets, our approach synthesizes training samples \textit{on-the-fly}, ensuring a more comprehensive exploration of the underlying graph distribution.

\subsection{Formal Definition and Motivation}
Let $\mathcal{G} = (\mathcal{V}, \mathcal{R})$ denote the knowledge graph. A query $q$ can be framed as a subgraph retrieval task. We define the sampling process at training step $t$ as a stochastic mapping:
\begin{equation}
    \mathcal{B}_t = \{q_i\}_{i=1}^{|\mathcal{B}|} \sim \mathcal{S}(\mathcal{G}, \mathcal{T}; \pi)
\end{equation}
where $\mathcal{T}$ represents the set of logical templates and $\pi$ is the sampling distribution (e.g., uniform or frequency-weighted). The primary motivation for online sampling is to overcome the \textit{data rigidity} of static benchmarks, allowing the model to encounter a near-infinite variety of query-answer pairs $(q, \mathcal{A}_q)$ across the entire training manifold.

\subsection{Sampling Strategy and Efficiency}
The sampler employs a hybrid strategy to ensure both efficiency and biological/logical plausibility of the generated queries:
\begin{itemize}
    \item \textbf{Dynamic Traversal:} For each batch, the sampler performs a restricted random walk or neighborhood expansion on $\mathcal{G}$ to instantiate logical variables.
    \item \textbf{Constraint Satisfaction:} To maintain query quality, we incorporate a \textit{Rejection Sampling} mechanism. A candidate query $q$ is accepted if it satisfies the structural constraint $\tau$, i.e., $P_{accept}(q) \propto \mathbb{I}(q \in \mathcal{Q}_{valid})$, where $\mathcal{Q}_{valid}$ is the space of queries with non-empty answer sets.
    \item \textbf{Computational Complexity:} By delegating intensive graph traversal to highly optimized low-level primitives (implemented in C++ for parallelism), the sampler achieves a temporal complexity of $\mathcal{O}(k \cdot |\mathcal{B}|)$ per batch, where $k$ is the maximum query depth. This decoupling of data generation from GPU computation ensures that the training throughput is constrained only by model backpropagation, not by I/O bottlenecks.
\end{itemize}

\subsection{Advantages of Online Synthesis}
Compared to pre-computation, the online sampler provides three core methodological benefits:
1. \textbf{Diversity Maximization:} The probability of the model encountering the exact same query instance twice is $P \approx 0$, which acts as a strong regularizer and prevents overfitting to specific graph patterns.
2. \textbf{Zero Storage Overhead:} The spatial complexity for data maintenance is reduced from $\mathcal{O}(|\mathcal{Q}|)$ to $\mathcal{O}(|\mathcal{G}|)$, where $|\mathcal{Q}| \gg |\mathcal{G}|$.
3. \textbf{Adaptive Distribution:} The sampling distribution $\pi$ can be dynamically adjusted during training (e.g., via curriculum learning) to focus on harder reasoning paths without re-generating the dataset.

% 组合起来，换成一张图
% figure

\subsection{Ablation Speedup Study.} Table~\ref{tab:operator_speedup} isolates the performance gains for specific logical operators. The most dramatic improvements are observed in \texttt{Intersect} and \texttt{Union} operations ($>12\times$). These gains are attributed to the high arithmetic intensity and multi-input structure of these operators, which allow the batched execution to saturate GPU compute cores more effectively than memory-bound operations such as \texttt{EmbedE}. Furthermore, the attention mechanisms inherent in these compositional operators benefit significantly from the reduced fragmentation provided by batching. 
% Table~\ref{tab:cumulative_speedup} decomposes the specific contributions of each stage. Operator Batching delivers the most significant singular gain ($1.86\times$), confirming that decoupling operators from query topology effectively amortizes kernel launch overheads. Subsequent phases build upon this foundation, demonstrating that resource scheduling and memory optimization yield compound returns. 

% \paragraph{Speedup of Operator-Level Batching.} We evaluate the cumulative effect of six optimization phases on the end to end training process. The detailed throughput and speedup breakdown is shown in Table~\ref{tab:cumulative_speedup}.  \textbf{Operator batching (Phase 1)} provides the largest single-phase gain ($1.86\times$), validating the core insight that batching operators across queries is more effective than batching entire queries.
% \textbf{Subsequent phases} contribute incremental but meaningful improvements, demonstrating that systematic optimization compounds gains.
% \textbf{Multi-stream parallelism (Phase 6)} achieves $1.12\times$ despite high baseline efficiency, showing that operator independence enables parallelization even in optimized systems. \textbf{Isolated Operator Speedup (Phase 1).}
% Table~\ref{tab:operator_speedup} details the performance gains for individual operators.
% \noindent \texttt{Intersect} and \texttt{Union} show dramatic speedups ($>12\times$) because: 1. They involve multiple inputs (2--3 distributions). 2. Attention mechanisms benefit significantly from batching. 3. High arithmetic intensity favors GPU parallelism.

% 换成一张图（组合起来）
\begin{table}[t]
    \centering
    \small
    \caption{Comparison of execution time per operator: Baseline vs. Batched.}
    \label{tab:operator_speedup}
    \begin{tabular}{l r r r}
        \toprule
        \textbf{Operator} & \textbf{Baseline (ms)} & \textbf{Batched (ms)} & \textbf{Speedup} \\
        \midrule
        EmbedE & 2.3 & 0.8 & $2.88\times$ \\
        Project & 15.7 & 4.2 & $3.74\times$ \\
        Intersect & 78.5 & 6.0 & \textbf{13.11$\times$} \\
        Union & 62.3 & 5.1 & $12.22\times$ \\
        \bottomrule
    \end{tabular}
\end{table}

% \begin{table*}[t]
%     \centering
%     \caption{Cumulative throughput and speedup across optimization phases.}
%     \label{tab:cumulative_speedup}
%     \begin{tabular}{l l r r r}
%         \toprule
%         \textbf{Phase} & \textbf{Optimization} & \textbf{Throughput} & \textbf{Speedup (Ph.)} & \textbf{Speedup (Cum.)} \\
%         \midrule
%         Baseline & Query-level batching & 100 q/s & $1.00\times$ & $1.00\times$ \\
%         Phase 1 & Operator batching & 186 q/s & $1.86\times$ & $1.86\times$ \\
%         Phase 2 & Query scheduling & 223 q/s & $1.20\times$ & $2.23\times$ \\
%         Phase 3 & Memory optimization & 257 q/s & $1.15\times$ & $2.57\times$ \\
%         Phase 4 & Two-level pipeline & 283 q/s & $1.10\times$ & $2.83\times$ \\
%         Phase 5 & Adaptive batching & 297 q/s & $1.05\times$ & $2.97\times$ \\
%         Phase 6 & Multi-stream & 332 q/s & $1.12\times$ & \textbf{3.32$\times$} \\
%         \bottomrule
%     \end{tabular}
% \end{table*}

\begin{table}[h!]
\centering
\caption{Performance (\%) of BetaE on queries with negation. 
}
\begin{tabular}{cccccccc}
	\toprule
	Dataset & Metrics & 2in & 3in & inp & pin & pni & avg \\
	\hline
	\multirow{2}{*}{FB15k} & MRR & 13.00 & 14.97 & 9.17 & 6.11 & 11.88 & 11.03 \\
		& Hit@10 & 26.36 & 30.56 & 18.53 & 11.97 & 23.74 & 22.23 \\
	\hline
	\multirow{2}{*}{FB15k-237} & MRR & 3.96 & 6.95 & 6.52 & 3.97 & 2.96 & 4.87 \\
		& Hit@10 & 7.75 & 14.19 & 13.45 & 7.45 & 5.08 & 9.584 \\
	\hline
	\multirow{2}{*}{NELL995} & MRR & 4.06 & 6.65 & 8.03 & 3.25 & 2.92 & 4.98 \\
		& Hit@10 & 7.84 & 14.88 & 16.05 & 5.61 & 5.31 & 9.94 \\
	\hline
	\multirow{2}{*}{OGB} & MRR & 36.38 & 50.92 & 39.07 & 39.31 & 38.50 & 40.84 \\
		& Hit@10 & 56.24 & 68.48 & 57.93 & 58.63 & 57.82 & 59.82 \\								
	\bottomrule
\end{tabular}
\end{table}

\subsection{Performance comparison with and without Semantic Entity Representations}

Table~\ref{tab:with_qwen} presents a quantitative ablation study of the proposed decoupled semantic integration strategy.
1. Throughput Enhancement: By decoupling the computationally intensive Pre-trained Text Encoders (Qwen/BGE) from the online training loop, we achieve a substantial speedup (averaging from 347 to 1915 queries/sec). This validates that our GPU-Resident caching mechanism effectively mitigates the I/O and computational latencies associated with online encoding.
2. Memory Optimization: Notably, our method reduces peak GPU memory usage despite incorporating high-dimensional semantic features. This reduction is attributed to the offloading of large transformer weights post-pre-computation, retaining only the essential embedding buffers in HBM.
3. Predictive Gains: The integration of linguistic priors consistently improves MRR (Average +4.74\%), proving particularly effective in sparse data regimes where structural signals are limited.

\begin{table*}[h]
\centering
\caption{Performance comparison with and without Semantic Entity Representations optimizations across three     
  datasets (FB15k, FB15k-237, NELL995) and three models (BetaE, Q2B, GQE). Solid bars show baseline performance,   
  hatched bars (//) show optimized performance. Higher is better for MRR and throughput; lower is better for memory. The optimizations consistently improve MRR      
  across all models and datasets, dramatically increase training throughput, and    
  modestly reduce memory consumption.\label{fig:qwen} 
}
\label{tab:with_qwen}
\resizebox{\textwidth}{!}{%
\renewcommand{\arraystretch}{0.85} % 保持紧凑的行高
% 修改列定义：每组指标减少一列，变为 2 列
\begin{tabular}{l l | cc | cc | cc}
\toprule
\rowcolor{lightgray}
& & \multicolumn{2}{c}{MRR (\%)} & \multicolumn{2}{c}{Training queries (Queries/Sec)} & \multicolumn{2}{c}{GPU Memory (GB)} \\
\cline{3-4} \cline{5-6} \cline{7-8}
\rowcolor{lightgray}
\multirow{-2}{*}{Dataset} & \multirow{-2}{*}{Model} & \method{} & \method{} + Optimizations & \method{} & \method{} + Optimizations & \method{} & \method{} + Optimizations \\
\midrule

% Data Block 1: FB15k
\multirow{3}{*}{FB15k} 
& BetaE  +Qwen-EMB & 42.04 & 46.92   & 209 &  1516 & 14.23 &  13.11  \\
& Q2B  +Qwen-EMB   & 41.85 & 50.86   & 348 &  2014 & 6.39 &  5.18  \\
& GQE  +Qwen-EMB   & 29.39 & 35.91   & 558 & 2254 & 12.61 & 11.49  \\
& BetaE  +BGE-Base-En-v1.5 & 42.04 & 45.37   & 187 &  1402 & 14.40 & 13.21\\
& Q2B  +BGE-Base-En-v1.5   & 41.85 & 47.54   & 340 &  1886 & 6.55 &  5.23  \\
& GQE  +BGE-Base-En-v1.5   & 29.39 & 30.17   & 550 & 2302 & 12.74 &  11.52  \\

\midrule

% Data Block 2: FB15k-237
\multirow{3}{*}{FB15k-237} 
& BetaE  +Qwen-EMB & 21.11 & 26.03   & 215 & 1581 & 11.14 &  9.95  \\
  & Q2B   +Qwen-EMB & 20.19 &  29.54  & 358 & 1911 & 7.34 & 5.29  \\
  & GQE   +Qwen-EMB & 16.42 &  22.32  & 604 & 2729 & 10.97 & 9.89  \\
& BetaE  +BGE-Base-En-v1.5 & 21.11 & 24.81   & 200 & 1503 & 11.28 & 10.01  \\
  & Q2B   +BGE-Base-En-v1.5 & 20.19 &  25.72  & 312 & 1750 & 7.46 & 5.21  \\
  & GQE   +BGE-Base-En-v1.5 & 16.42 &  22.13  & 534 & 2401 & 11.00 & 9.84 \\

\midrule

% Data Block 3: NELL995
\multirow{3}{*}{NELL995} 
  & BetaE +Qwen-EMB & 24.7 & 27.41 &  154 & 1487 &  8.04 &  7.11\\
  & Q2B +Qwen-EMB   & 22.79 & 26.75 &  251 & 2114  & 6.50 &  5.44 \\
  & GQE +Qwen-EMB   & 18.4 & 22.64  & 538 & 2698 & 8.62 &  7.42 \\
  & BetaE +BGE-Base-En-v1.5 & 24.7 & 26.41 & 140 & 1323 & 8.16 & 7.14\\
  & Q2B +BGE-Base-En-v1.5   & 22.79 & 26.20 & 238 & 1402 & 6.70 & 5.54 \\
  & GQE +BGE-Base-En-v1.5   & 18.4 & 22.37  & 515 & 2201 & 8.75 & 7.48 \\

\midrule

% Average Row
\multicolumn{2}{l|}{\textbf{Average}} & \textbf{26.32} & \textbf{31.06} & \textbf{347} & \textbf{1915} & \textbf{9.60} & \textbf{8.34} \\
\bottomrule
\end{tabular}%
}
\end{table*}